\PassOptionsToPackage{table,svgnames,dvipsnames}{xcolor}

\documentclass[]{bytedance_seed}

\usepackage[toc,page,header]{appendix}

\usepackage{minitoc}
\usepackage{cleveref}
\usepackage{subcaption}
\usepackage{booktabs}
\usepackage{graphicx}
\usepackage{CJKutf8}
\usepackage{xcolor}
\usepackage{tabularx}
\usepackage{float}
\usepackage{caption}
\usepackage{listings}
\usepackage{tcolorbox}
\tcbuselibrary{listingsutf8}

\definecolor{datasetprompt}{HTML}{EFE8E3}
\definecolor{datasetpromptheader}{HTML}{CFC6C1}

\newtcolorbox{datasetpromptbox}[1][]{
  enhanced, breakable,
  top=0.3em,bottom=0.3em,left=0.5em,right=0.5em,
  toptitle=0.3em,bottomtitle=0.2em,boxsep=0pt,
  colframe=datasetpromptheader,
  colback=datasetprompt,
  boxrule=0.5pt,
  width=\linewidth,
  title={\footnotesize #1}
}

\usepackage{natbib}

\usepackage{latexsym}
\usepackage{url}
\usepackage{amsmath}
\usepackage{amssymb}
\usepackage{amsfonts}
\usepackage{mathtools}
\usepackage{amsthm}
\usepackage[utf8]{inputenc}
\usepackage{microtype}

\usepackage{booktabs}
\usepackage{multirow}
\usepackage{makecell}
\usepackage{paralist}
\usepackage{xspace}
\usepackage{pifont}
\usepackage{colortbl}
\usepackage{adjustbox}
\usepackage{array}
\usepackage{enumitem}
\usepackage{verbatim}

\usepackage{hyperref}
\usepackage{caption}

\hypersetup{
    colorlinks,
    linkcolor={blue!80!black},
    citecolor={blue!80!black},
}

\title{Thinking with Drafting: Optical Decompression via Logical Reconstruction}

\author[1,2,5,*]{Jingxuan~Wei}
\author[1,2,*]{Honghao~He}
\author[1,2]{Caijun~Jia}
\author[3]{Siyuan~Li}
\author[1,2]{Zheng~Sun}
\author[1,2]{Yuhang~Xu}
\author[3]{Yuanyuan~Lin}
\author[1,2]{Linzhuang~Sun}
\author[3]{Yuchen~Wu}
\author[1,2]{Bihui~Yu}
\author[3,\dagger]{Xiangxiang~Zhang}
\author[4,\dagger]{Cheng~Tan}

\affiliation[1]{Shenyang Institute of Computing Technology, Chinese Academy of Sciences}
\affiliation[2]{University of Chinese Academy of Sciences\quad $^{3}$ByteDance\quad $^{4}$Westlake University}
\affiliation[5]{Key Laboratory of Computing Power Network and Information Security, Ministry of Education, Shandong Computer Science Center (National Supercomputer Center in Jinan), Qilu University of Technology (Shandong Academy of Sciences)}

\contribution[*]{Equal contribution}
\contribution[\dagger]{Corresponding authors}

\date{January 5, 2026}
\correspondence{tancheng@pjlab.org.cn, zhangxiangxiang.zxx@bytedance.com}

\abstract{
Existing multimodal large language models have achieved high-fidelity visual perception and exploratory visual generation. However, a precision paradox persists in complex reasoning tasks: optical perception systems transcribe symbols without capturing logical topology, while pixel-based generative models produce visual artifacts lacking mathematical exactness. To bridge this gap, we propose that reasoning over visual inputs be reconceptualized as \textit{optical decompression}---the process of reconstructing latent logical structures from compressed visual tokens. Guided by the axiom that \textit{Parsing is Reasoning}, we introduce Thinking with Drafting (TwD), which utilizes a minimalist Domain-Specific Language (DSL) as a grounding intermediate representation. Unlike standard approaches that hallucinate answers directly, TwD forces the model to draft its mental model into executable code, rendering deterministic visual proofs for self-verification. To validate this, we present VisAlg, a visual algebra benchmark. Experiments demonstrate that TwD serve as a superior cognitive scaffold. Our work establishes a closed-loop system where visual generation acts not as a creative output but as a logical verifier, offering a generalizable path for visual reasoning.

}

\begin{document}

\maketitle


\section{Introduction}

Recent advances in multimodal large language models (MLLMs) mark a decisive shift in artificial intelligence from passive perception toward active cognitive interaction~\cite{huang2023language,alayrac2022flamingo,xue2024xgen,liu2023visual}. On the input side, \textbf{optical character recognition (OCR)} systems have undergone a dramatic evolution. Modern approaches---exemplified by large-scale vision-language models trained for document understanding---are now capable of faithfully transcribing complex visual artifacts~\cite{kim2022ocr,wang2024mineru,cui2025paddleocr,tang2023unifying}, including dense text, structured layouts, tables, and mathematical formulas. This progress effectively realizes what may be termed contextual optical compression~\cite{wei2025deepseek}: rich visual documents are compressed into high-fidelity internal representations, enabling machines to read with unprecedented accuracy.

\begin{figure}[!htbp]
    \centering
    \includegraphics[width=\linewidth]{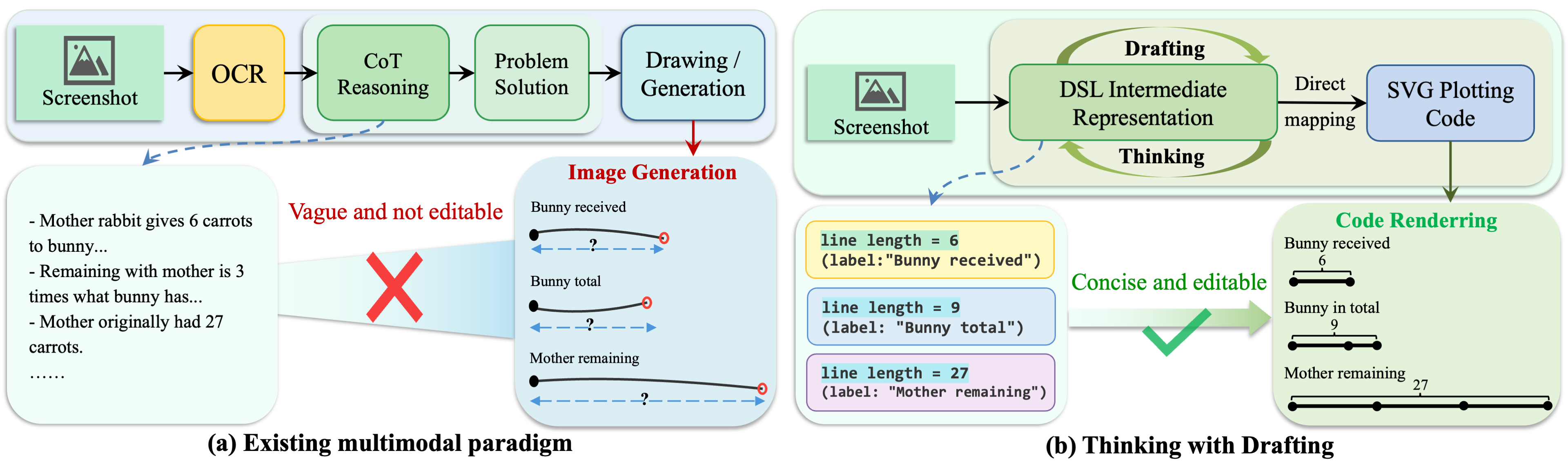}
    \vspace{-1.5em}
    \caption{Illustration of paradigms. (a) Existing multimodal paradigms treat image understanding, textual reasoning, and visual generation as disconnected tasks. (b) Thinking with Drafting (TwD) reframes visual reasoning as logical reconstruction into a minimalist DSL.}
    \vspace{-4mm}
    \label{fig:intro}
\end{figure}

Concurrently, progress on the output side has given rise to a complementary paradigm often referred to as \textbf{Thinking with images}~\cite{su2025thinking,chern2025thinking}. Rather than relying solely on a textual chain-of-thought, recent models increasingly generate visual artifacts like diagrams, sketches, or intermediate images as part of the reasoning process. By externalizing cognition into visual form, these methods aim to mirror a fundamental aspect of human problem solving, where drawing and visualization serve as tools for thought~\cite{zheng2025deepeyes,qiao2025v}. Taken together, these trends suggest that modern systems are approaching a read--draw loop~\cite{lu2023chameleon,shen2023hugginggpt}, with perception supplying faithful inputs and generation enabling visualized intermediate states.

Despite this apparent completeness, a critical gap remains when these systems are applied to tasks requiring strict logical precision. This gap manifests as a precision paradox. On the one hand, OCR systems excel at transcription: they can reliably extract symbols, numbers, and text spans from images. However, transcription alone does not capture logical topology. A numeral such as ``123'' may represent a total, a difference, or a constraint, depending on context. While the perceptual signal is high-fidelity, the relational semantics remain implicit and unstructured. OCR systems are designed to recognize symbols, represent the logical relations that govern them. On the other hand, visual generation models optimize perceptual plausibility rather than logical validity~\cite{yao2022react}. They can generate images that resemble diagrams or mathematical constructions without guaranteeing that the underlying relations are exact. A generated line segment may appear longer than another, yet fail to satisfy a precise quantitative ratio.

To bridge this divide, we argue that reasoning over visual inputs must be reconceptualized as a process of \textit{optical decompression}~\cite{schick2023toolformer,hsu2023ns3d}. If OCR compresses the visual world into perceptual tokens, then reasoning is the act of reconstructing the latent logical structure encoded within those tokens. From this perspective, understanding does not hinge on producing fluent textual explanations, but on recovering an explicit, executable representation of entities, relations, and constraints. This leads to our central axiom: \textbf{Parsing is Reasoning}. True comprehension arises only when a model can translate ambiguous natural language and visual cues into a structured form.

We materialize this philosophy through the \textbf{Thinking with Drafting (TwD)} paradigm. Taking the Singapore bar model---a canonical representation of visual algebra---as our primary testbed, we introduce a minimalist geometric DSL (Domain-Specific Language). This DSL occupies a unique strategic niche: it serves as an intermediary between the ambiguity of natural language, the syntactic noise of general-purpose code, and the rigidity of geometric axioms. This DSL is designed for interoperability; it can be compiled into GeoGebra scripts for mathematical validation or SVG code for visual rendering. The generated draft serves not merely as a visualization, but as a deterministic visual verifier, enabling the system to detect logical conflicts and self-correct. Within TwD, drafting is not treated as a final output but as a \emph{deterministic visual verifier}, enabling a closed logical--visual loop in which reconstruction, verification, and correction are tightly coupled.

\section{Related Work}

\subsection{Optical Perception}

Recent advancements in Optical Character Recognition (OCR)~\cite{wang2024mineru,li2025monkeyocr,cui2025paddleocr} and Vision Language Models (VLMs)~\cite{hurst2024gpt,comanici2025gemini,bai2025qwen2,yang2025qwen3} have fundamentally transformed the landscape of document understanding. Traditional approaches have evolved to recover high-fidelity text content while preserving complex contextual structures such as layouts, tables, and formulas~\cite{zhang2025docr,chumachenko2025nvidia}. Notably, recent works like DeepSeek-OCR~\cite{wei2025deepseek} demonstrate the feasibility of contexts optical compression, proving that pixels can serve as an efficient compression medium for textual information.

However, optical perception alone is insufficient for tasks that require rigorous logical consistency like mathematical problem solving~\cite{gupta2023visual,suris2023vipergpt,lu2021inter}. Current unstructured outputs may capture the document's visual syntax but neglect its underlying logic, leaving entities and quantitative relations implicit and ungrounded. We argue that reliable reasoning requires a shift from transcription accuracy to logical reconstruction. Unlike standard perception tasks, our approach transforms raw perception into a verifiable intermediate representation, thereby enabling the \textit{Thinking with Drafting} paradigm to operate on grounded logical structures.

\subsection{Visual Reasoning}

While optical perception digitizes the input, reasoning requires manipulating digitized concepts to derive solutions. The dominant paradigm relies on LLMs to perform reasoning via textual generation, exemplified by Chain-of-Thought (CoT)~\cite{wei2022chain,kojima2022large} and Program-of-Thought (PoT)~\cite{chenprogram,gao2023pal}. These methods decompose complex problems into step-by-step deductions or executable code snippets. Conversely, vision-centric approaches attempt to solve reasoning tasks directly in the pixel space~\cite{chen2025learning,yang2025machine,zhang2023multimodal}. Recent works such as Vision-ARC~\cite{hu2025arc} demonstrate that certain abstract reasoning tasks are more naturally formulated as image-to-image translation problems.

Despite their efficacy, these models often struggle with \textit{semantic grounding}---specifically, translating complex natural language constraints into geometric artifacts. We propose \textit{Thinking with Drafting} to bridge the gap between implicit semantic thought and explicit visual verification. By parsing visual text into a structured intermediate representation, the model drafts its understanding into a rule-constrained canvas. It creates an \textit{optical decompression} loop: implicit logical relations are decompressed into explicit visual structures.

\begin{figure}[!htbp]
    \centering
    \includegraphics[width=1.0\linewidth]{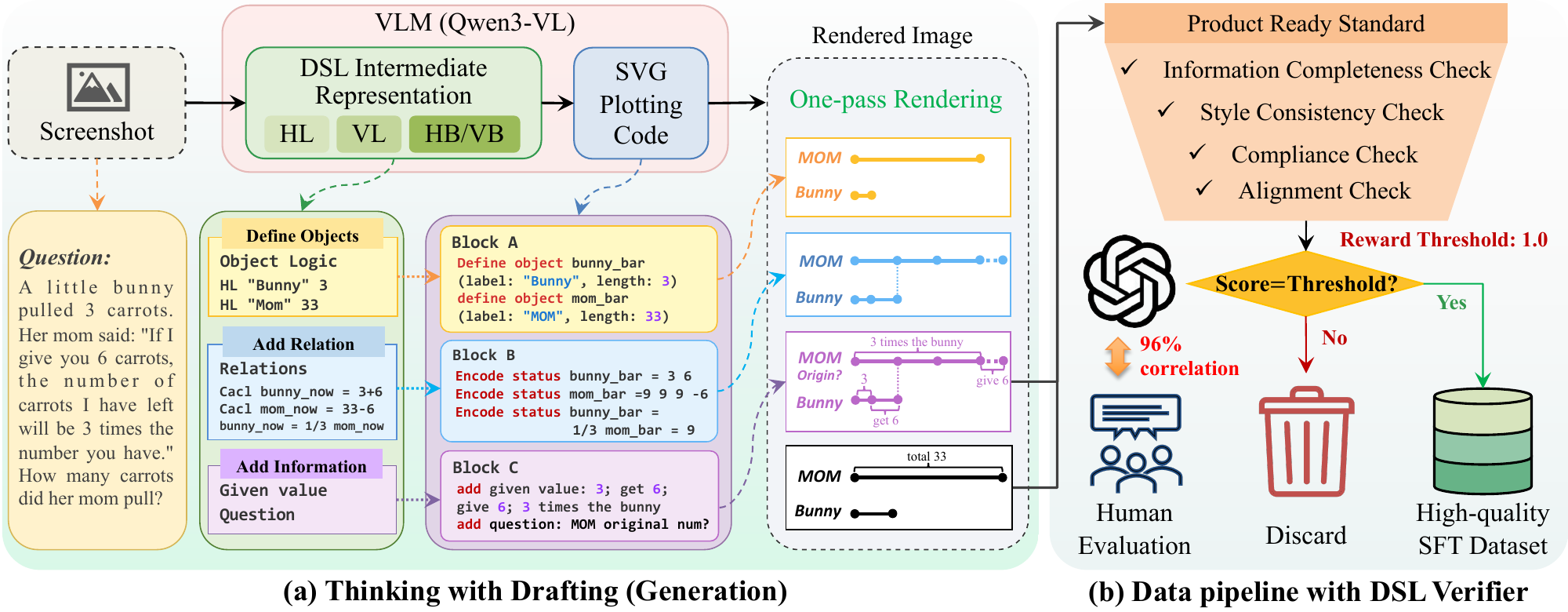}
    \vspace{-1.75em}
    \caption{Overview of Thinking with Drafting framework.
    (a) Optical decompression generates a Logic Graphic DSL from visual input and OCR, comprising entity, relational, and aggregation primitives. (b) A verifier scores samples by syntactic validity, visual completeness, and logical consistency, retaining high-quality data for training and discarding the rest to ensure topological and geometric correctness.}
    \vspace{-2mm}
    \label{fig:framework}
\end{figure}

\section{Method}

\subsection{Preliminaries}

We consider the problem of multimodal mathematical reasoning where a model is presented with a visual input $\mathcal{I}$ (containing visual text, layout, and geometry) and a natural language query $\mathcal{Q}$. The objective is to derive a correct final answer $a \in \mathcal{A}$. Unlike standard end-to-end approaches that map $(\mathcal{I}, \mathcal{Q}) \rightarrow a$ directly, we formalize \textit{Thinking with Drafting} as a multi-stage iterative generation process involving a structured intermediate representation. Let $\mathcal{T}$ denotes the space of unstructured natural language and $\mathcal{S}$ denotes the space of DSL, which represents geometric and logical constraints. We define the reasoning process $P_\theta$, parameterized by a MLLM, as a probabilistic mapping from perception to logical reconstruction.

To underscore the theoretical distinctiveness of TwD, we contrast our formulation with three dominant paradigms: text-only CoT, thinking with images, and traditional OCR.

\paragraph{Text-only CoT}
Standard Multimodal CoT approaches rely exclusively on the linguistic space $\mathcal{T}$ to bridge the input and output:
\begin{equation}
     \hat{t}_{cot}\sim P_\theta(t,\mathcal{I}, \mathcal{Q},),\; \hat{a} \sim P_\theta(a | \mathcal{I}, \mathcal{Q}, \hat{t}_{cot}),
\end{equation}
where $\hat{t}_{cot} \in \mathcal{T}$ is a linear sequence of tokens. The fundamental limitation of CoT is that natural language is ambiguous and lacks strict geometric constraints. In contrast, our DSL space $\mathcal{S}$ enforces logical rigidity; a defined entity in $\mathcal{S}$ must satisfy explicit geometric rules, acting as a regularizer for the reasoning process.

\paragraph{Thinking with Images}
Emerging ``Thinking with Images'' paradigms utilize a generative model to produce an intermediate image $\hat{\mathcal{I}}_{\text{gen}}$:
\begin{equation}
    \hat{a} \sim P_\theta(a \mid \mathcal{I}, \mathcal{Q}, \hat{\mathcal{I}}_{\text{gen}})
\end{equation}
While $\hat{\mathcal{I}}_{\text{gen}}$ provides visual feedback, it operates in the pixel space, which suffers from stochastic imprecision. A model may generate a diagram that \textit{perceptually} plausible but \textit{mathematically} inaccurate. TwD, conversely, employs programmatic drafting. Our intermediate representation $\hat{s}$ is symbolic code. The rendered output is mathematically exact, ensuring reliable verification.

\paragraph{OCR} focuses on transcription fidelity, mapping the visual input to a sequence of characters:
\begin{equation}
    \text{Seq} \sim P_\theta(\text{Seq} \mid \mathcal{I}),
\end{equation}
OCR addresses the question ``\textit{What is written?}'', whereas TwD addresses ``\textit{What does it mean?}''. OCR extracts the \textit{syntax} but leaves the \textit{semantics} implicit. TwD performs logical reconstruction, upgrading the task from transcription to parsing. By mapping $\mathcal{I} \to \mathcal{S}$, we explicitly capture the logical topology that OCR ignores, thereby converting raw pixels into actionable reasoning primitives.

\subsection{The Logic Graphic DSL}

To instantiate the principle that \textit{Parsing is Reasoning}, we formally define the structure of our DSL space, $\mathcal{S}$. A statement $s\in\mathcal{S}$ is not a sequence of natural language tokens, but a structured composition of atomic reasoning primitives. Unlike general-purpose plotting languages that prioritize pixel-level control, the grammar of $\mathcal{S}$ is designed to abstract away rendering redundancies and expose the bare logical topology of the problem. The DSL consists of three fundamental operator categories:

\paragraph{Entity Primitives (\textsc{HL})}
These represent the physical quantities or objects from the input $\mathcal{I}$ as horizontal line segments. A key innovation in our design is the status-aware segmentation. We define a segment sequence vector $\mathbf{v} = [v_1, v_2,...,v_n]$ where $|v_i|$ denotes length. Crucially, we utilize the sign of $v_i$ to encode \textit{existential status}: $v_i > 0$ renders a solid line (existing quantity), while $v_i < 0$ renders a dashed line (process quantity, e.g., subtracted part or hypothetical extension). This allows the model $P_\theta$ to generate a compact representation for complex change models.

\paragraph{Relational Primitives (\textsc{VL})}
In bar models, logic is primarily defined by geometric alignment. The Vertical Line (\textsc{VL}) operator explicitly encodes relational equality between horizontal entities. Parameterized by an explicit $x$-coordinate and row indices, it functions as an equality constraint, asserting that specified segments coincide at a shared value. This compels the model to perform \textit{alignment reasoning}, identifying shared semantic boundaries rather than treating coordinates as independent variables.

\paragraph{Aggregation Primitives (\textsc{HB/VB})}
To ground abstract arithmetic operations into geometry, we employ Horizontal (\textsc{HB}) and Vertical (\textsc{VB}) Braces. An \textsc{HB} operator encapsulates a part-whole relationship within a single entity, while a \textsc{VB} represents summation or comparison across multiple entities.

\subsection{Topological Abstraction and Rendering}
A major bottleneck in generating visual code is the high entropy of continuous coordinate spaces. To mitigate this, we introduce a Topological Abstraction layer that decouples logical reasoning from metric rendering.

\paragraph{Virtual Grid System}
We map the continuous canvas $\mathbb{R}^2$ to a discrete logic space $\mathbb{Z}^2$. We define a virtual grid where the $y$-axis is discretized into logical rows and the $x$-axis is governed by relative offsets rather than absolute pixels. The model generates code relative to this grid. For instance, creating a new entity involves assigning it to a new row\_id rather than calculating a pixel offset. It ensures layout invariance: the model focuses solely on the logical ordering and grouping of entities.

\paragraph{Deterministic Rendering}
The mapping from a syntactically correct DSL statement to a visual verification image $\mathcal{V}$ is executed by a deterministic rendering engine: $\mathcal{V} = \textrm{Render}(s)$. We introduce common topological patterns into semantic macros. For example, a \textit{comparison pattern} macro automatically generates the difference brace and alignment lines when the model detects a more than relation. These macros ensure that correct logical parsing always yields a visually canonical diagram.

\subsection{Thinking with Drafting}

Building upon the structured space $\mathcal{S}$ and the deterministic renderer, we instantiate the TwD framework in Figure~\ref{fig:framework} as a sequential generation-verification process.

\paragraph{Optical Decompression via Logical Parsing}
In the first stage, the model acts as a parser. It perceives the raw input $\mathcal{I}$ and attempts to decompress the implicit logical topology into an explicit structural draft. This yields a preliminary textual explanation $\hat{t}$ and an initial draft $\hat{s}$:
\begin{equation}
    (\hat{t},\hat{s}) \sim P_\theta (t,s | \mathcal{I}, \mathcal{Q}),
\end{equation}
Crucially, the generation of $\hat{s}$ is not a single step, but a step-by-step decomposition of the problem. It embodies our axiom that \textit{Parsing is Reasoning}: the generation of $s$ forces the model to resolve ambiguities in $\mathcal{I}$ into discrete logical atoms.

\paragraph{Drafting and DSL-Conditioned Inference}
The generated hypothesis $\hat{s}$ is passed to the rendering engine to produce the verification drafting image $\mathcal{V}$. It provides an explicit visual proof of the model's internal reasoning for human verification. The model derives the solution conditioned on the structured draft $\hat{s}$ it constructed. Unlike standard Chain-of-Thought which relies on ambiguous natural language, the drafting $\hat{s}$ acts as a logical context:

In the second stage, the model utilizes the output from the previous stage as a ``drafting context.'' The initial draft $\hat{s}_1$ acts as an externalized cognitive scaffold, allowing the model to inspect its own reasoning. The model generates a refined explanation $\hat{t}_2$, a completed DSL $\hat{s}_2$, and the final answer $\hat{a}$:
\begin{equation}
    \hat{a} \sim P_\theta (t,s,a| \mathcal{I}, \mathcal{Q}, \hat{t}, \hat{s}).
\end{equation}
By grounding the reasoning in $\hat{s}$, the calculations are guided by the explicit topology defined in the draft. The TwD paradigm thus posits that the \textit{act of constructing} the draft is the reasoning engine itself, ensuring the final answer is a derivative of a verified logical structure.

\section{Dataset}

\begin{figure}[!htbp]
    \centering
    \includegraphics[width=\linewidth]{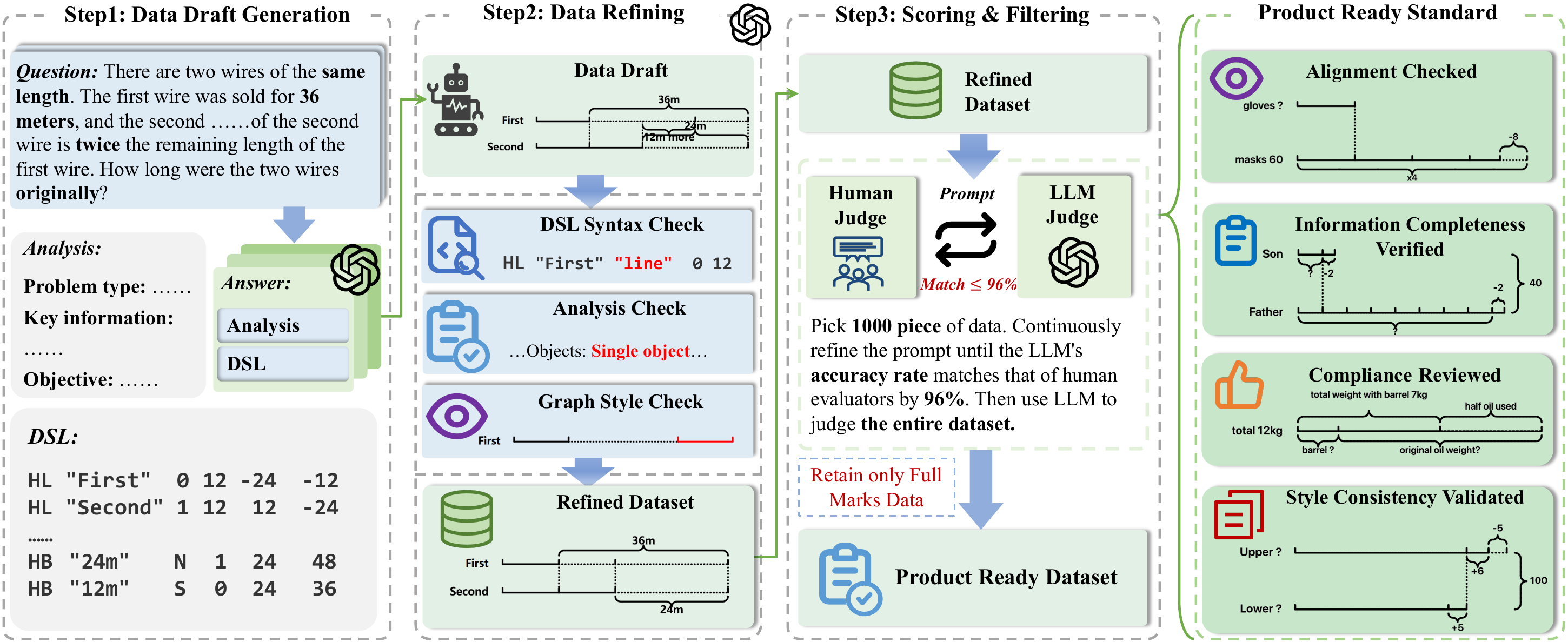}
    \caption{The benchmark data construction pipeline of VisAlg.}
    \label{fig:pipeline}
\end{figure}

We introduce VisAlg, a benchmark for evaluating logic-aware visual reasoning by assessing whether a system can recover the explicit logical topology underlying visual algebra problems through \textit{optical decompression}. Each instance pairs a image of a natural language algebra problem with a structured intermediate representation, an executable bar-model DSL that defines the ground-truth logical parse. VisAlg is constructed through a multi-stage pipeline, as shown in Figure~\ref{fig:pipeline}.

\subsection{Dataset Construction}
\label{sec:dataset_construction}

\noindent\textbf{Drafting data generation.}
We collect \(15{,}000\) bar-model word problems from public datasets and websites, covering common visual algebra patterns. For each problem, we prompt Gemini-2.5-Pro~\cite{comanici2025gemini} to produce a synchronized draft with two components. The first component is a textual analysis that explicitly parses the problem schema, and the second component is a program written in our DSL. The detailed prompt is provided in Appendix~\ref{sec:appendix_draft_prompt}.

\noindent\textbf{Data refining.}
Initial drafts frequently fail to meet verifiability requirements. We therefore introduce a checklist refinement stage in which the model revises each draft through three sequential checks:
(1) \textit{Syntax check}, ensuring the grammar is correct and executable;
(2) \textit{Analysis check}, verifying that all objects, quantities, relations, and targets identified in the analysis are consistently instantiated;
(3) \textit{Style check}, enforcing canonical bar-model layout conventions such as boundary placement and cross-row alignment.
All corrected instances are stored.
The detailed prompt is in Appendix~\ref{sec:appendix_refine_prompt}.

\noindent\textbf{Scoring and filtering.}
We employ an LLM-based judge calibrated with expert evaluations to filter the dataset. A domain expert scores \(1{,}000\) instances using a fixed rubric, and the judge prompt is iteratively refined until achieving \(96\%\) agreement. The calibrated judge is then applied to the full dataset, retaining only full-score instances, resulting in \(11{,}372\) product-ready instances.
Details are provided in Appendix~\ref{sec:appendix_score_prompt} and~\ref{sec:appendix_human_eval}.

\noindent\textbf{Product ready.}
The final filtering enforces four criteria: geometric alignment, semantic completeness, representational compliance, and stylistic consistency.
(1) \textit{Alignment}: Horizontal bracket endpoints must coincide with boundary coordinates defined by cumulative segment lengths; vertical links must align with these boundaries across spanned rows.
(2) \textit{Completeness}: All stated quantities, relations, and targets must appear explicitly in labels; unknowns may be denoted by ``?''.
(3) \textit{Compliance}: Vertical brackets represent only multi-object aggregates, and vertical links are allowed solely at cross-row shared partition points.
(4) \textit{Consistency}: Transfers follow a paired $-t/+t$ pattern across two rows; post-transfer equality is indicated by a shared boundary via a vertical link.

\subsection{Dataset Analysis}
\label{sec:dataset_analysis}

\begin{figure}[!htbp]
  \centering
  \begin{minipage}[t]{0.46\linewidth}
    \centering
    \vspace{0pt}
    \includegraphics[width=\linewidth]{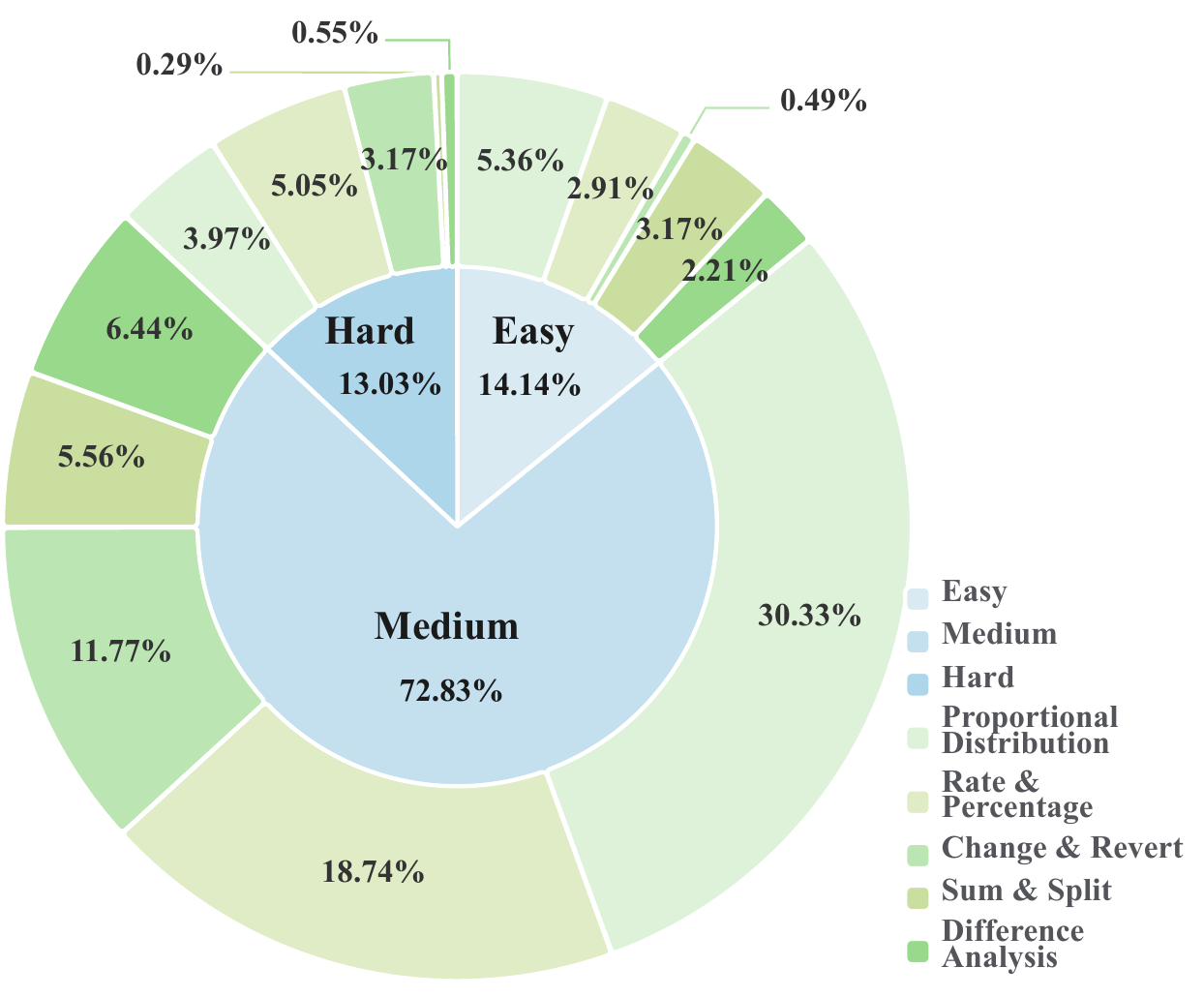}
    \captionof{figure}{Difficulty and schema composition in VisAlg.}
    \label{fig:data_distribution}
  \end{minipage}\hfill
  \begin{minipage}[t]{0.52\linewidth}
    \centering
    \vspace{0pt}
    \captionof{table}{Corpus-level statistics of VisAlg.}
    \label{tab:visalg_stats}
    \small
    \setlength{\tabcolsep}{3mm}
    \begin{tabular}{lrr}
      \toprule
      & \textbf{Train} & \textbf{Test} \\
      \midrule
      \multicolumn{3}{l}{\textbf{Problem schemas}} \\
      \hline
      Proportional Distribution & 4{,}265 & 245 \\
      Rate \& Percentage        & 2{,}771 & 265 \\
      Change \& Revert          & 1{,}635 & 119 \\
      Difference Analysis       &   905 & 141 \\
      Sum \& Split              &   854 & 172 \\
      \midrule
      \multicolumn{3}{l}{\textbf{Difficulty levels}} \\
      \hline
      Easy   & 1{,}400 & 208 \\
      Medium & 7{,}602 & 680 \\
      Hard   & 1{,}428 & 54 \\
      \midrule
      \multicolumn{3}{l}{\textbf{Operation length}} \\
      \hline
      $\leq$ 3 operations & 1{,}834 & 163 \\
      4 operations        & 2{,}490 & 279 \\
      5 operations        & 2{,}693 & 296 \\
      6 operations        & 1{,}612 & 115 \\
      7 operations        & 1{,}002 & 56 \\
      $\geq$ 8 operations &   799 & 33 \\
      \midrule
      \textbf{Total instances} & \textbf{10{,}430} & \textbf{942} \\
      \bottomrule
    \end{tabular}
  \end{minipage}
\end{figure}

\noindent\textbf{Category.} VisAlg focuses on optical decompression in bar-model reasoning, emphasizing recovery of logical topology over surface symbol transcription. We analyze five canonical schemas: proportional distribution, rate \& percentage, change \& revert, sum \& split, and difference analysis. Figure~\ref{fig:data_distribution} shows the joint composition of difficulty (inner ring) and schema (outer ring).
As summarized in Table~\ref{tab:visalg_stats}, proportional distribution and rate \& percentage form the two dominant schema groups.
In terms of difficulty, medium accounts for 72.9\% of instances, with easy (13.4\%) and hard (13.7\%) providing balanced coverage of both basic parsing and constraint-dense cases.

\noindent\textbf{Reasoning depth.}
Logical complexity is measured by the number of bar-model operations needed for reconstruction. As shown in Table~\ref{tab:visalg_stats}, only 17.6\% of training instances require three or fewer operations, while most fall in the four--six range. A non-trivial 7.7\% require eight or more operations, reflecting long dependency chains and multi-step reasoning.

\noindent\textbf{Scale and split consistency.}
The benchmark comprises 10,430 training and 942 test instances, with additional curated splits for fine-tuning, preference optimization, and evaluation. The test set mirrors the training distribution in schema and difficulty, ensuring evaluation emphasizes structural generalization rather than distributional shift.

\subsection{Evaluation Metrics}
\label{sec:evaluation_metrics}

\noindent\textbf{Objective metrics.}
Consistency is evaluated at both code and image levels. Code similarity is measured using BLEU, ROUGE-L, and chrF, with chrF as the primary metric due to its robustness to mixed symbols, numbers, and text in the DSL. Image similarity is assessed using PSNR, SSIM, and LPIPS, with SSIM prioritized for its sensitivity to structural topology and edge continuity.

\noindent\textbf{Subjective metrics via LLM-as-judge.}
An LLM-based verifier scores outputs on five dimensions: structural alignment, information coverage, numerical consistency, semantic compliance, and answer leakage. Each is rated in $[0,1]$, with the final subjective score given by their mean.

\noindent\textbf{Main score.}
Main results are reported using a composite score:
$\mathrm{Score} = \frac{1}{3}\bigl(\mathrm{chrF} + \mathrm{SSIM} + \mathrm{LLM}_{\text{judge}}\bigr)$,
which jointly reflects code-level consistency, image-level structural fidelity, and semantic normative correctness.

\noindent\textbf{Human evaluation.}
We additionally conduct a human evaluation of DSL quality; the criteria and protocol are provided in Appendix~\ref{app:human_eval_criteria}.

\section{Experiment}

\begin{table}[!htbp]
  \centering
  \vspace{-2mm}
  \caption{Main results on VisAlg. Align: structural alignment; Cover: information coverage; Num: numerical consistency; Norm: semantic compliance; Leak: answer leakage. Detailed descriptions of them are in Appendix~\ref{app:eval_dim}.}
  \vspace{-2mm}
  \label{tab:main_results}
\setlength{\tabcolsep}{1.0mm}
  \resizebox{\linewidth}{!}{
  \begin{tabular}{l|ccc|ccc|cccccc|c}
    \toprule
    \multicolumn{1}{l|}{\multirow{2}[4]{*}{Model}} &
    \multicolumn{3}{c|}{Code Similarity} &
    \multicolumn{3}{c|}{Image Similarity} &
    \multicolumn{6}{c|}{Verification Scores} &
    \multirow{2}[4]{*}{Overall} \\
\cmidrule{2-13}
     & BLEU & ROUGE-L & chrF & LPIPS & SSIM & PSNR
    & Align & Cover & Num & Norm & Leak & Avg. & \\
    \midrule
    \textcolor[rgb]{.122,.137,.161}{InternVL3-8B~\cite{zhu2025internvl3}} & 9.93 & 48.51 & 37.57 & 32.64 & 82.70 & 24.25 & 0.31 & 0.56 & 0.32 & 0.15 & 0.89 & 44.69 & 54.99 \\
    \textcolor[rgb]{.122,.137,.161}{InternVL2.5-8B~\cite{chen2024expanding}} & 9.12 & 46.38 & 48.41 & 51.10 & 58.97 & 17.22 & 0.32 & 0.44 & 0.36 & 0.14 & 0.68 & 38.73 & 48.70 \\
    \textcolor[rgb]{.122,.137,.161}{Intern-S1-mini~\cite{bai2025intern}} & 8.41 & 36.47 & 22.68 & 59.52 & 49.32 & 14.65 & 0.57 & 0.26 & 0.87 & 0.36 & 0.97 & 60.39 & 44.13 \\
    \textcolor[rgb]{.122,.137,.161}{Mimo-VL-7B-RL~\cite{li2025xiaomi}} & 10.36 & 46.17 & 33.43 & 79.47 & 25.87 & 7.75 & 0.38 & 0.48 & 0.76 & 0.23 & 0.85 & 54.05 & 37.78 \\
    \textcolor[rgb]{.122,.137,.161}{Qwen3-VL-8B~\cite{bai2025qwen3vltechnicalreport}} & 6.65 & 39.04 & 23.94 & 83.69 & 20.10 & 0.00 & 0.60 & 0.16 & 0.84 & 0.29 & \cellcolor[rgb]{.851,.953,.992}\textbf{1.00} & 57.80 & 33.95 \\
    \midrule
    Gemini-3-Pro~\cite{team2023gemini} & \cellcolor[rgb]{.851,.961,.839}30.18 & \cellcolor[rgb]{.851,.961,.839}59.06 & \cellcolor[rgb]{.851,.961,.839}57.53 & 18.23 & \cellcolor[rgb]{.851,.961,.839}90.36 & \cellcolor[rgb]{.851,.961,.839}27.32 & \cellcolor[rgb]{.851,.953,.992}\textbf{0.97} & \cellcolor[rgb]{.851,.961,.839}0.95 & 0.94 & \cellcolor[rgb]{.851,.953,.992}\textbf{0.78} & 0.96 & \cellcolor[rgb]{.851,.953,.992}\textbf{91.98} & \cellcolor[rgb]{.851,.961,.839}79.96 \\
    Gemini-2.5-Pro~\cite{comanici2025gemini} & 28.94 & 58.25 & 57.43 & \cellcolor[rgb]{.851,.961,.839}18.12 & 89.97 & 26.92 & \cellcolor[rgb]{.851,.961,.839}0.95 & 0.76 & \cellcolor[rgb]{.851,.953,.992}\textbf{0.99} & \cellcolor[rgb]{.851,.961,.839}0.73 & 0.32 & 74.97 & 74.12 \\
    Claude-4~\cite{Anthropic2025ClaudeSonnet4_5} & 28.66 & 58.54 & 57.17 & 18.16 & 89.97 & 26.88 & 0.94 & 0.74 & \cellcolor[rgb]{.851,.961,.839}0.99 & 0.72 & 0.30 & 73.71 & 73.62 \\
    GPT-5.1~\cite{achiam2023gpt} & 22.93 & 56.13 & 51.23 & 25.86 & 86.89 & 25.70 & 0.77 & 0.67 & 0.95 & 0.51 & 0.19 & 61.69 & 66.60 \\
    GPT-4o~\cite{hurst2024gpt} & 16.24 & 50.64 & 35.73 & 24.49 & 89.15 & 26.59 & 0.50 & 0.42 & 0.83 & 0.31 & 0.72 & 55.44 & 60.11 \\
    \midrule
    TwD (Ours) & \cellcolor[rgb]{.851,.953,.992}\textbf{48.23} & \cellcolor[rgb]{.851,.953,.992}\textbf{72.22} & \cellcolor[rgb]{.851,.953,.992}\textbf{68.29} & \cellcolor[rgb]{.851,.953,.992}\textbf{11.97} & \cellcolor[rgb]{.851,.953,.992}\textbf{93.68} & \cellcolor[rgb]{.851,.953,.992}\textbf{30.25} & 0.90 & \cellcolor[rgb]{.851,.953,.992}\textbf{0.96} & 0.70 & 0.73 & \cellcolor[rgb]{.851,.961,.839}1.00 & \cellcolor[rgb]{.851,.961,.839}85.91 & \cellcolor[rgb]{.851,.953,.992}\textbf{82.63} \\
    \bottomrule
  \end{tabular}
  }
  \vspace{-2mm}
\end{table}

\subsection{Experimental Setup}
\label{sec:experimental_setup}

We evaluate VisAlg against state-of-the-art MLLMs.
The proprietary models include GPT-5.1~\cite{achiam2023gpt}, GPT-4o~\cite{hurst2024gpt}, Claude-4~\cite{Anthropic2025ClaudeSonnet4_5}, Gemini-3~\cite{team2023gemini}, and Gemini-2.5-Pro~\cite{comanici2025gemini}, representing the current upper bound of general-purpose multimodal reasoning.
For open-weight baselines, we consider InternVL3-8B~\cite{zhu2025internvl3}, InternVL2.5-8B~\cite{chen2024expanding}, Intern-S1-mini~\cite{bai2025intern}, Mimo-VL-7B-RL~\cite{li2025xiaomi}, and Qwen3-VL-8B~\cite{bai2025qwen3vltechnicalreport}.
Our model is initialized from Qwen3-VL-8B and supervised fine-tuned on the training split, enabling parameter-efficient comparison with open-weight peers while treating proprietary models as upper bounds. SFT is conducted on a 8-GPU node with a visual token cap of 2,048 and a maximum sequence length of 5,128. We train for 2 epochs using a learning rate of $5\times10^{-6}$ and a warmup ratio of 0.05.

\subsection{Main Results}
\label{sec:main_results}

Table~\ref{tab:main_results} reports the main results on VisAlg across code similarity, image similarity, and verifier-based evaluation.
Our model, initialized from Qwen3-VL-8B and supervised on VisAlg, achieves the highest overall score of 82.63, surpassing all open-weight baselines and outperforming the strongest proprietary models, including Gemini-3-Pro~\cite{team2023gemini} (79.96) and Gemini-2.5-Pro~\cite{comanici2025gemini} (74.12). This highlights the importance of explicit supervision on logic reconstruction for verifiable bar-model reasoning.
A clear performance gap is observed between open-weight and proprietary systems.
Open-weight models such as InternVL3-8B~\cite{zhu2025internvl3}, InternVL2.5-8B~\cite{chen2024expanding}, Intern-S1-mini~\cite{bai2025intern}, Mimo-VL-7B-RL~\cite{li2025xiaomi}, and Qwen3-VL-8B~\cite{bai2025qwen3vltechnicalreport} score below 55, with weaknesses in code fidelity and diagram reconstruction, indicating difficulty in generating syntactically valid and topologically consistent DSL programs without task-specific alignment.

Our gains primarily arise from improved structural fidelity. The model leads in code and diagram alignment, achieves strong information coverage, and avoids answer leakage. Relative to top proprietary models, the remaining gap is mainly in numerical consistency, while structural legality and semantic completeness are largely preserved.

\subsection{Results by Visual Algebra Schema}
\label{sec:schema_results}
\begin{figure}[!htbp]
  \centering
  \includegraphics[width=\linewidth]{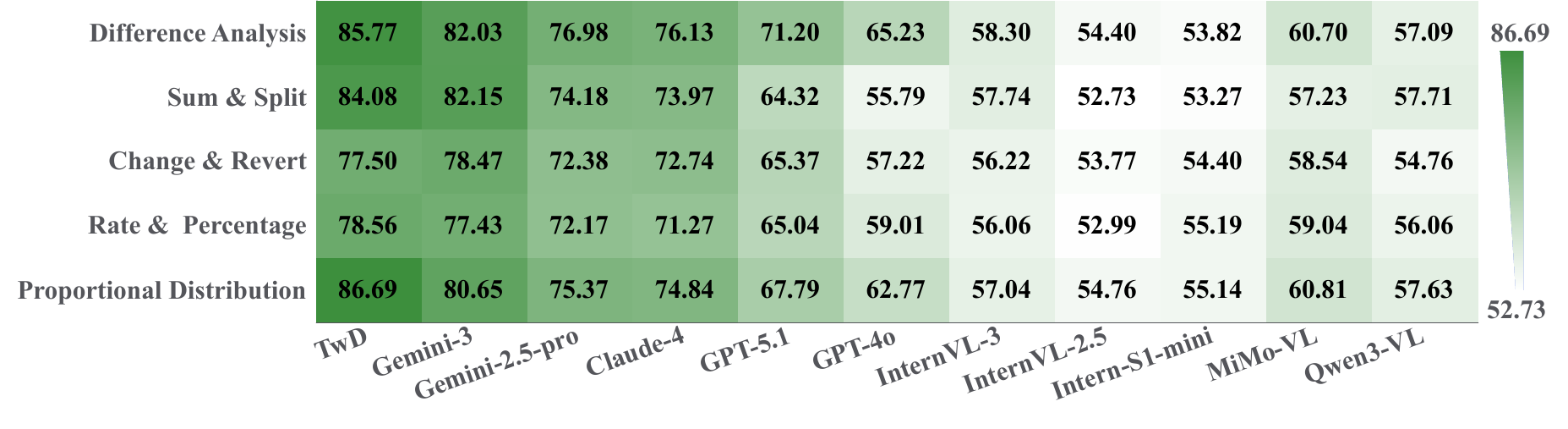}
  \vspace{-9mm}
  \caption{Schema-wise performance comparison across five visual algebra problem types.}
  \vspace{-2mm}
  \label{fig:schema_heatmap}
\end{figure}

Figure~\ref{fig:schema_heatmap} reports schema-wise performance across five visual algebra types.
Our TwD consistently achieve competitive performance compared to both open-weight and proprietary baselines across all schemas.
The gains are most pronounced on structure-intensive schemas such as \emph{proportional distribution} and \emph{difference analysis}, where accurate multi-segment decomposition and boundary-aligned comparison are critical. While proprietary models achieve competitive results, their performance varies noticeably across schemas. In contrast, TwD remains uniformly strong across problem types, supporting the claim that optical decompression benefits from explicit, verifiable logic.

\subsection{Alignment with Human Expert}
\label{sec:human_alignment}

\begin{figure}[!htbp]
  \centering
  \begin{minipage}[t]{0.48\linewidth}
    \centering
    \vspace{0pt}
    \includegraphics[width=\linewidth]{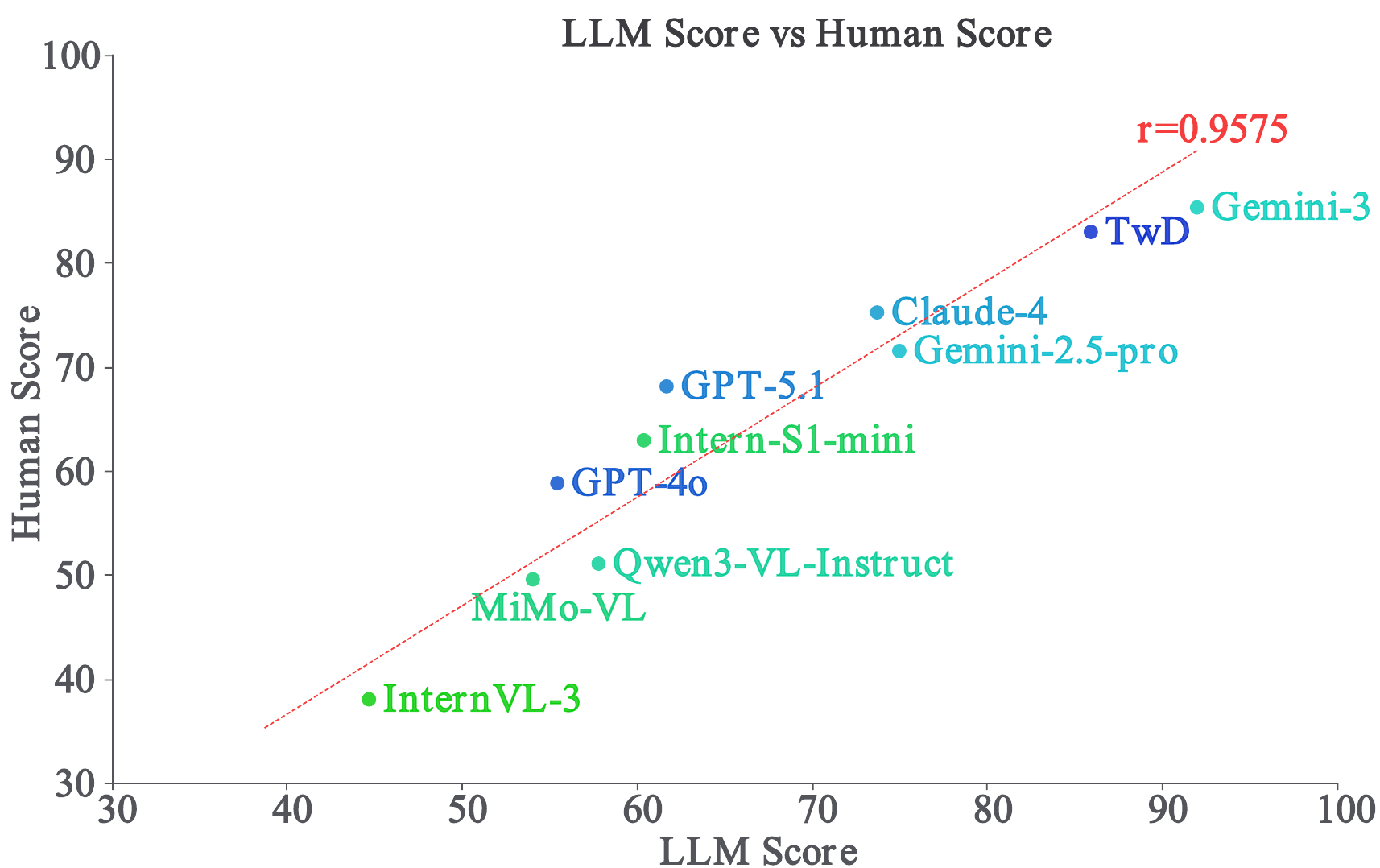}
    \captionof{figure}{Correlation between verifier-based VisAlg scores and human expert ratings.}
    \label{fig:human_llm_correlation}
  \end{minipage}\hfill
  \begin{minipage}[t]{0.48\linewidth}
    \centering
    \vspace{0pt}
    \includegraphics[width=\linewidth]{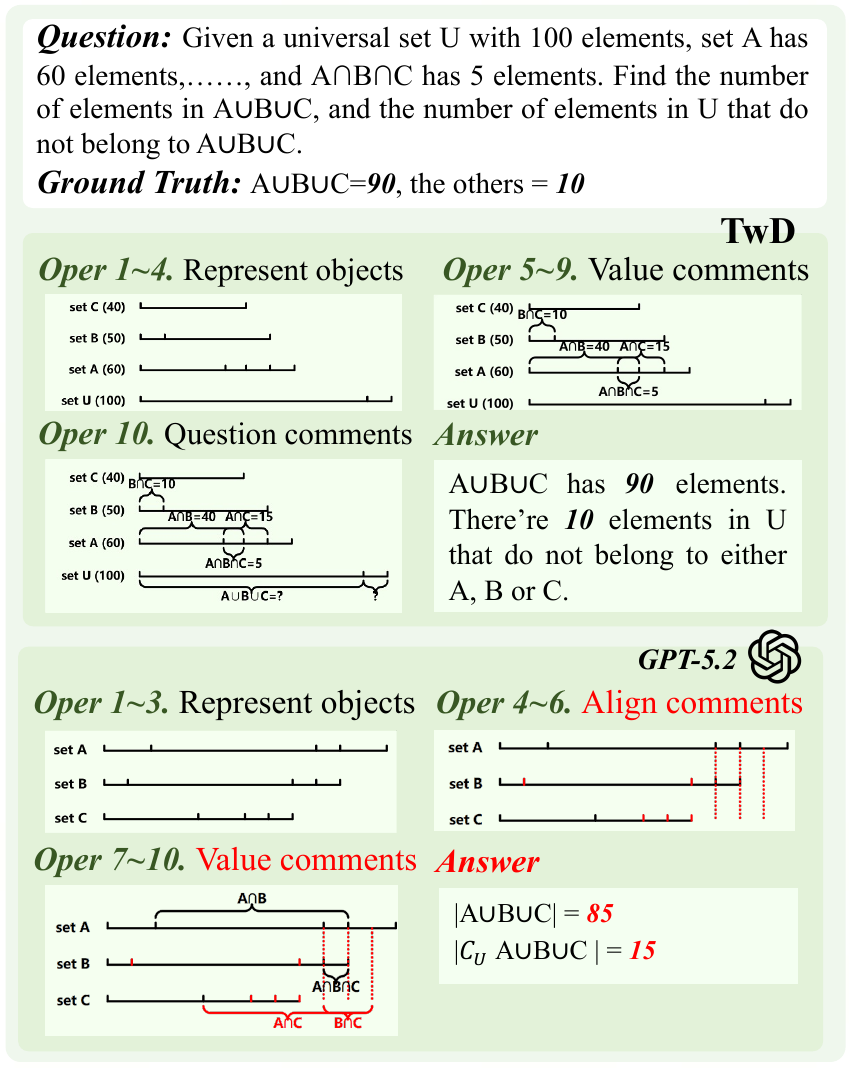}
    \captionof{figure}{Generalization to set-theoretic reasoning.}
    \label{fig:error_sets}
  \end{minipage}
\end{figure}

Figure~\ref{fig:human_llm_correlation} shows a strong correlation between expert human ratings and verifier-based VisAlg scores $r=0.9575$, validating the verifier as a reliable proxy for human judgment in visual algebra reasoning. Model rankings are largely preserved across the full performance range. TwD remains top-ranked under both evaluations, indicating that the reported gains reflect genuine improvements in structural correctness rather than metric artifacts.

\subsection{Generalize to Complex Logical Topology}
\label{sec:error_analysis}

We extend our evaluation to advanced set-theoretic reasoning tasks involving multi-set constraints. As shown in Figure~\ref{fig:error_sets}, these tasks require the model to manage high-order intersections and nested boolean boundaries. Frontier MLLMs like GPT-5~\cite{achiam2023gpt} often exhibit topological hallucination in this regime. While they may attempt to align segments visually, they fail to preserve the strict boolean logic of overlaps. The model cannot distinctively ground intersections $A\cap C$ and $A\cap B\cap C$, violating containment and alignment constraints and rendering the graphic unreadable and unverifiable.
This \textit{calculation--construction gap} highlights that correct arithmetic does not guarantee preservation of global structural invariants such as boundary legality and consistency.
TwD successfully decomposes the abstract set problem into sequential geometric operations. By explicitly rendering the atomic intersections, TwD effectively visualizes the algebra of sets.
Additional case studies are provided in Appendix~\ref{app:more_error}.

\section{Conclusion}
\label{sec:conclusion}

In this work, we addressed the precision paradox in multimodal reasoning, where systems achieve high perceptual fidelity yet fail to preserve rigorous logical topology. We formalized this challenge through the lens of \textit{optical decompression}, introducing VisAlg benchmark to evaluate whether models can reconstruct latent logical structures into verifiable artifacts. To bridge the gap between perception and reasoning, we established Thinking with Drafting paradigm that enforces structural invariants via a minimalist graphic DSL. Experiments demonstrate that that a compact 8B model, when equipped with the TwD cognitive scaffold, outperforms leading proprietary frontiers on visual algebra problems. By closing this loop, we show that explicit structural drafting acts as a necessary foundation for trustworthy multimodal intelligence.

\section*{Limitations}

The core limitation lies in the scope of structural representation: the DSL is intentionally designed around bar-model visual algebra, emphasizing linear topological relations to enable intuitive structural supervision. Extending this DSL to support broader classes of scientific diagrams remains an important direction for future research.

\clearpage

\bibliographystyle{plain}
\bibliography{main}

\clearpage
\beginappendix
\section{Additional Details for Dataset Construction}
\label{sec:appendix_dataset}

\subsection{Prompt for Data Draft Generation}
\label{sec:appendix_draft_prompt}

This subsection presents the prompt used in Step~1 (Data Draft Generation) of the VisAlg construction pipeline.
The prompt elicits a synchronized draft consisting of structured problem analysis, diagram planning under strict bar-model constraints, and an initial executable DSL program.
This stage establishes the logical and visual foundation for subsequent refinement and verification.
The complete prompt is provided in Figure~\ref{draft_generation_prompt}.

\begin{figure*}[!ht]
\centering
\begin{datasetpromptbox}[Step 1: Data Draft Generation Prompt]

\footnotesize
You are an expert instructor in bar-model reasoning.
You specialize in constructing bar-model diagrams using a structured \textbf{Graphic Intermediate Representation (DSL)} to support mathematical problem solving.

Given a math word problem and its associated image, \textbf{directly} produce the output in the following format.
Do \textbf{not} include any preamble, explanation, or commentary.

\textbf{Step 1: Problem Analysis}

Analyze the core structure of the problem using concise language.
Do not restate the problem text.
Output only the following four items:

(1) \textbf{Problem type}: e.g., sum--difference, proportional, transfer, comparison.

(2) \textbf{Objects}: specify whether the problem involves a single object or multiple independent objects.

(3) \textbf{Key information}: short phrases describing all given quantities and relations.

(4) \textbf{Query}: one sentence stating the required unknown.

\textbf{Step 2: Diagram Planning}

List the key considerations required to construct a correct and verifiable bar-model diagram:

(1) Whether a vertical bracket (VB) is permitted (VB is forbidden for single-object problems and allowed only for multi-object aggregation).

(2) Whether all \emph{given} information from the problem is explicitly labeled in the diagram under a \textbf{visible-text-only} policy (numerical segment lengths do not count as textual labels).

(3) Whether the queried quantity is explicitly marked in the diagram using interrogative naming (e.g., ``total ?'', ``how many ?'').

(4) The core operation type involved (reduction, increase, transfer, comparison) and its corresponding visual encoding.

(5) Whether vertical alignment lines (VL) are required to express cross-row shared boundaries or post-operation equality.

(6) Any special constraints, including strict prohibition of answer leakage in all labels.

\textbf{Step 3: Initial DSL Draft}

Perform the necessary internal reasoning to determine correct segment lengths, then output an initial DSL program.
Output \textbf{only} a clean DSL code block, without comments, explanations, or blank lines.

\texttt{```dsl}\\
\texttt{<DSL code only>}\\
\texttt{```}

\textbf{Output Format Requirements}

(1) Begin \textbf{directly} with ``Step 1: Problem Analysis''.

(2) Use the exact step titles shown above.

(3) Output only the three steps; do not generate any additional content.

\textbf{DSL Syntax}

(1) \texttt{HL ``name'' row l1 l2 \ldots}: horizontal bar composed of semantic subsegments; positive values denote solid segments, negative values denote dashed segments (absolute value as length).

(2) \texttt{VL x row0 row1}: vertical alignment line used only for cross-row shared boundaries.

(3) \texttt{HB ``label'' N|S row x0 x1}: horizontal bracket spanning interval $[x_0, x_1]$, placed above (N) or below (S) the bar.

(4) \texttt{VB ``label'' col row0 row1}: vertical bracket used \textbf{only} for aggregating multiple independent objects.

\textbf{Core Construction Rules (Strict)}

(1) All horizontal bars must be decomposed into semantic subsegments; drawing a single undivided bar is not allowed.

(2) Fractions and ratios must be represented via equal-length subsegments; multiplicative relations must be expressed by repeating equal segments, with the base quantity placed on the upper row.

(3) Reduction must be encoded as \texttt{(A - t)} followed by \texttt{(-t)}, with solid segments on the left and dashed segments on the right; negative segments denote \emph{only} removal, deficit, transfer-out, or unknown placeholders.

(4) Transfer must be represented as paired $-t/+t$ segments across rows; post-transfer equality must be marked with a shared VL at the corresponding boundary.

(5) Vertical alignment lines are permitted only for cross-row shared boundaries and must align exactly with segment boundaries on \emph{all} involved rows.

(6) Bracket endpoints must coincide exactly with segment boundaries; floating or misaligned brackets are invalid.

(7) Final numerical answers must not appear in any labels; computed values may appear only as segment lengths for rendering purposes.

(8) All given information from the problem statement must appear explicitly as textual labels in the diagram.

\textbf{Silent Self-Check}

Before outputting, internally verify alignment correctness, semantic compliance, information completeness, and non-leakage of final answers.

\end{datasetpromptbox}
\caption{Prompt used in Step~1 for generating structured analysis and the initial DSL draft during VisAlg dataset construction.}
\label{draft_generation_prompt}
\end{figure*}

\subsection{Prompt for Data Refining}
\label{sec:appendix_refine_prompt}

This subsection presents the prompt used in Step~2 of the VisAlg construction pipeline.
Given an initial problem analysis and a draft DSL generated in the previous stage, this prompt instructs the model to perform checklist-driven verification and conditional refinement.
The objective is to determine whether the draft is \emph{product-ready}, and if not, to apply minimal, targeted corrections.
The full checklist-driven refinement prompt is shown in Figure~\ref{fig:refine_prompt}.

\begin{figure*}[t]
\centering
\begin{datasetpromptbox}[Checklist-Driven Refinement Prompt]

\footnotesize
You are an expert instructor in bar-model reasoning.
You specialize in auditing and refining bar-model diagrams expressed in a structured \textbf{Graphic Intermediate Representation (DSL)}.

You are given a math word problem with its image, together with an existing analysis and an initial DSL draft.
Your task is to \textbf{verify the draft using a checklist} and \textbf{revise it only if violations are found}.
Do \textbf{not} include any preamble or commentary.

\textbf{Checklist-Based Verification}

Select \textbf{2--4} verification items that are most relevant to the given problem.
The following two checks are \textbf{mandatory} and must always be included.
Each check must be marked as either \texttt{[PASS]} or \texttt{[FAIL]}.

(1) \textbf{Alignment}:
verify that all bracket endpoints coincide with bar-segment boundaries or shared alignment positions, and that any vertical alignment markers lie on valid bar segments.

(2) \textbf{Information completeness}:
verify that all original quantities and relations stated in the problem are explicitly labeled in the diagram, and that the queried quantity is clearly indicated.

In addition, select applicable checks from the following categories.

(3) \textbf{Norm compliance}:
verify that reductions follow the left-solid/right-dashed convention;
comparisons use right-side dashed segments;
and transfers are encoded as paired subtraction and addition across rows.

(4) \textbf{Style consistency}:
verify that multiplicative structures place the base quantity on the upper row;
and that bracket placement prioritizes upper positioning, resolving overlaps by span length.

\textbf{Refinement Decision}

State in one sentence whether the DSL draft satisfies all selected checks.

If any check is marked \texttt{[FAIL]}, briefly describe the violations and output a corrected DSL program.
If all checks pass, state that the draft can be used without modification.

\textbf{Final Answer Generation}

After the refinement decision, continue \emph{directly} with the final solution generation.
The solution must include, in order:
(1) a one-sentence identification of the problem type;
(2) a concise reasoning outline with formulas written in \texttt{\$...\$};
(3) a clean DSL code block representing the final diagram;
(4) step-by-step computations; and
(5) the final numerical answer.

\textbf{Output Constraints}

(1) Begin output directly with the checklist-based verification section.\\
(2) Use bold text only for section headers; do not use numbered steps or section markers.\\
(3) Do not repeat the problem statement.\\
(4) Do not include any content outside the specified structure.

\textbf{DSL Semantics}

All DSL syntax and construction rules strictly follow those defined in the draft-generation stage, including semantic subsegment decomposition, VB usage restrictions, reduction and transfer encoding, alignment constraints, and the prohibition of answer leakage.

\end{datasetpromptbox}
\caption{Prompt used for checklist-driven verification and conditional refinement of initial DSL drafts during VisAlg dataset construction.}
\label{fig:refine_prompt}
\end{figure*}

\subsection{Prompt for Scoring and Filtering}
\label{sec:appendix_score_prompt}

This subsection presents the prompt used in the final stage of the VisAlg construction pipeline.
Given a refined DSL draft produced after checklist-driven revision, this prompt instructs an LLM-based verifier to perform strict, criteria-based scoring.
Only instances receiving a full score are retained as \emph{product-ready} samples in the final dataset.
The scoring prompt used for LLM-based verification is shown in Figure~\ref{fig:score_prompt}.

\begin{figure*}[t]
\centering
\begin{datasetpromptbox}[Scoring Prompt for DSL Verification]

\footnotesize
You are an expert instructor in bar-model reasoning.
You are familiar with evaluating bar-model diagrams expressed in a structured \textbf{Graphic Intermediate Representation (DSL)}.

Your task is to \textbf{score the given DSL code} based solely on the \textbf{problem statement} and the \textbf{provided solution (including DSL code)}.
You must \textbf{not} modify, complete, or reinterpret the DSL code.
Do not introduce any information beyond what is explicitly provided.

\textbf{Evaluation Scope}

All judgments must follow the \emph{visible-text-only} principle:
only quoted strings in \texttt{HL} names and \texttt{HB}/\texttt{VB} labels are considered visible annotations.
Numeric segment lengths and coordinates are not treated as textual information.

\textbf{Scoring Criteria}

Evaluate the DSL code according to the following checklist.
For each item, output either \texttt{[PASS]} or \texttt{[FAIL]}, together with a brief justification.

\textbf{Critical Criteria (Fail Any $\Rightarrow$ Score = 0.0)}

(1) \textbf{Alignment correctness}: all bracket endpoints align with bar-segment boundaries; all vertical alignment markers lie on valid shared boundaries.

(2) \textbf{Information completeness}: all given quantities and required unknowns from the problem statement are explicitly annotated using visible text.

(3) \textbf{Numerical consistency}: all segment lengths are numerically self-consistent with the problem logic, without arbitrary scaling or unexplained values.

(4) \textbf{Transfer correctness}: transfer operations must be represented by paired subtraction and addition across rows, with appropriate alignment markers when required.

(5) \textbf{Answer leakage}: no final answer values appear in any visible annotations.

(6) \textbf{VB/VL usage}: vertical brackets are used only for multi-object aggregation; vertical alignment markers appear only at shared cross-row boundaries.

\textbf{Non-Critical Criteria}

The following criteria affect the score only if all critical criteria pass:

(7) \textbf{Reduction conventions}: reduction and deficit relations follow the left-solid/right-dashed convention.

(8) \textbf{Multiplicative structure}: multiplicative relations are expressed using repeated equal-length segments, with the base quantity placed on the upper row.

(9) \textbf{Semantic decomposition}: horizontal bars are decomposed into semantically meaningful subsegments rather than drawn as undivided totals.

(10) \textbf{Label conciseness}: visible annotations are concise, non-redundant, and free of embedded calculations.

\textbf{Scoring Rule}

If any critical criterion is marked \texttt{[FAIL]}, the final score is \textbf{0.0}.
Otherwise, the base score is \textbf{1.0}, with a penalty of \textbf{0.1} deducted for each failed non-critical criterion.

\textbf{Output Format}

First output a section titled \texttt{[Scoring Rationale]}, listing each criterion with its pass/fail status and justification.
Then output a single line:

\begin{verbatim}
[Final Score]: <float between 0.0 and 1.0>
\end{verbatim}

Do not output JSON or any additional formatting.

\end{datasetpromptbox}
\caption{Prompt used for strict LLM-based scoring and filtering of refined DSL drafts in VisAlg.}
\label{fig:score_prompt}
\end{figure*}

\subsection{Human Expert Evaluation Criteria}
\label{sec:appendix_human_eval}

In addition to automated LLM-based verification, we perform human expert screening on all refined instances.
Human evaluators assess each DSL using the same zero-tolerance philosophy, serving as the final gatekeeper for dataset inclusion.

An instance is accepted into the final dataset \textbf{only if all of the following conditions are satisfied}:

(1) \textbf{Numerical validity}: all bar-segment lengths correspond to valid quantities in the correct solution process, without arbitrary scaling or distortion.

(2) \textbf{Information sufficiency}: the rendered diagram alone, based on visible annotations, is sufficient to solve the problem without consulting the original text.

(3) \textbf{Alignment accuracy}: all brackets and alignment markers precisely coincide with valid segment boundaries.

(4) \textbf{Semantic fidelity}: the diagram correctly encodes object relationships described in the natural language problem.

(5) \textbf{Format compliance}: all constructions adhere strictly to the prescribed DSL conventions for reduction, transfer, multiplicative relations, and alignment.

Only instances that satisfy all five criteria are retained as \emph{product-ready} samples in VisAlg.

\section{Human Evaluation Criteria for Evaluation Metrics}
\label{app:human_eval_criteria}

\subsection{Evaluation Target and Boundary Conditions}
The human assessment is strictly limited to the \emph{given} DSL output. Reviewers must evaluate whether the DSL expresses the problem's key quantities and relationships in a \emph{norm-compliant}, \emph{non-leaking}, and \emph{structurally self-consistent} manner, such that a reader can reliably reconstruct the intended structure and carry out a correct derivation from the diagram.

To minimize subjective preference, prior belief, and post-hoc ``mental correction,'' the evaluation is conducted under the following boundary conditions:
(i) reviewers must not introduce any extra information beyond what is explicitly encoded in the DSL;
(ii) reviewers must not modify the output in any form, including adding segments, changing numeric values, renaming labels, or reformatting the program;
(iii) the assessment does not consider the writing quality, fluency, or style of any accompanying natural-language solution.

\subsection{Review Protocol and Evidence-Driven Practice}
A structured, evidence-driven expert review protocol is adopted to maximize objectivity and reproducibility.
Three domain experts with backgrounds in mathematics education and diagram-oriented coding are recruited.
Each sample is rated independently by at least two reviewers; disagreements spanning two or more score levels are resolved by a third reviewer via arbitration.

To discourage intuition-based judgments, each assigned rating must be accompanied by \emph{minimal sufficient evidence} that is directly verifiable from the DSL.
Typical evidence includes: an HB endpoint failing to coincide with an HL segment boundary; a VL failing to align with a cross-row critical boundary; a quoted label explicitly containing the final numeric answer to the queried quantity (answer leakage); or arithmetic constraints that cannot hold under the implied sum--difference or transfer relations.
Evidence logs enable third-party auditing without reliance on reviewer-specific interpretation.

\subsection{Evaluation Dimensions}
\label{app:eval_dim}
DSL quality is characterized along five dimensions that jointly determine usability:

\paragraph{Structural Alignment.}
Whether HB endpoints and VL coordinates strictly coincide with HL segment boundaries, reflecting geometric legality and representational precision.

\paragraph{Information Coverage.}
Whether all key givens and the queried unknown are explicitly marked or clearly represented based \emph{only} on visible textual labels (i.e., quoted strings).
Numeric segment lengths alone do not count as textual labels.
This dimension measures whether the intended problem structure is recoverable from the diagram content.

\paragraph{Numerical Consistency.}
Whether the segment lengths satisfy the intended arithmetic constraints (sum, difference, increase/decrease, transfer amount).
Systematic errors such as uniform scaling artifacts or the use of numerous uninterpretable numbers are treated as violations.

\paragraph{Semantic Conformity.}
Whether the DSL follows task-specific construction conventions, including: reduction encoded as solid-left and dashed-right segments; transfer encoded as paired $-t/+t$ segments with equal magnitude across rows; multiplicative relations expressed via repeated equal-length subsegments with the base quantity emphasized; non-abusive use of VL/VB; and semantically motivated HL decomposition by problem type.

\paragraph{Answer Leakage.}
A hard constraint assessed solely from visible textual labels: if the final numeric answer to the queried quantity appears explicitly in quoted labels, the output is considered leaking.

\subsection{Overall Rating Scale}
An overall five-level score is assigned based on the five dimensions above:

\begin{itemize}
  \item \textbf{5 (Excellent):} No leakage; strict structural alignment; complete information coverage; numerically consistent relations; and clear, norm-compliant semantic decomposition.
  \item \textbf{4 (Good):} Overall correct and readable; minor non-critical imperfections (e.g., slight alignment or labeling issues) that do not hinder understanding or derivation.
  \item \textbf{3 (Acceptable):} Still usable for problem solving but requires frequent reference to the problem statement to resolve ambiguities; partial missing labels, coarse decomposition, or localized norm violations may be present.
  \item \textbf{2 (Poor):} High risk of misinterpretation due to unreliable alignment, missing critical information, multiple semantic violations, or strained numerical relations, making stable derivation difficult.
  \item \textbf{1 (Unacceptable):} Fatal violations, including answer leakage, uniform scaling artifacts, fundamentally invalid sum--difference or transfer relations, large-scale alignment failures, or incorrect core semantic structure.
\end{itemize}

\subsection{Leakage as a Dominant Violation}
Answer leakage is treated as the most destructive violation because it breaks the boundary that the diagram should encode structure rather than disclose the solution.
Accordingly, once leakage is confirmed (from quoted labels), the output is rated as unacceptable and the evidence must be recorded explicitly.

\section{Case Studies on Visual Algebra Schemas}
\label{app:case_studies}

We present one representative example for each of the five visual algebra schemas in VisAlg.
These cases illustrate how \textit{Thinking with Drafting} (TwD) operationalizes \emph{optical decompression}: it converts abstract textual constraints into an explicit and spatially aligned \emph{DSL}, so that the problem can be solved directly from the rendered structure without relying on implicit, unverified reasoning.

\begin{figure}[ht]
  \centering
  \includegraphics[height=0.78\textheight,keepaspectratio]{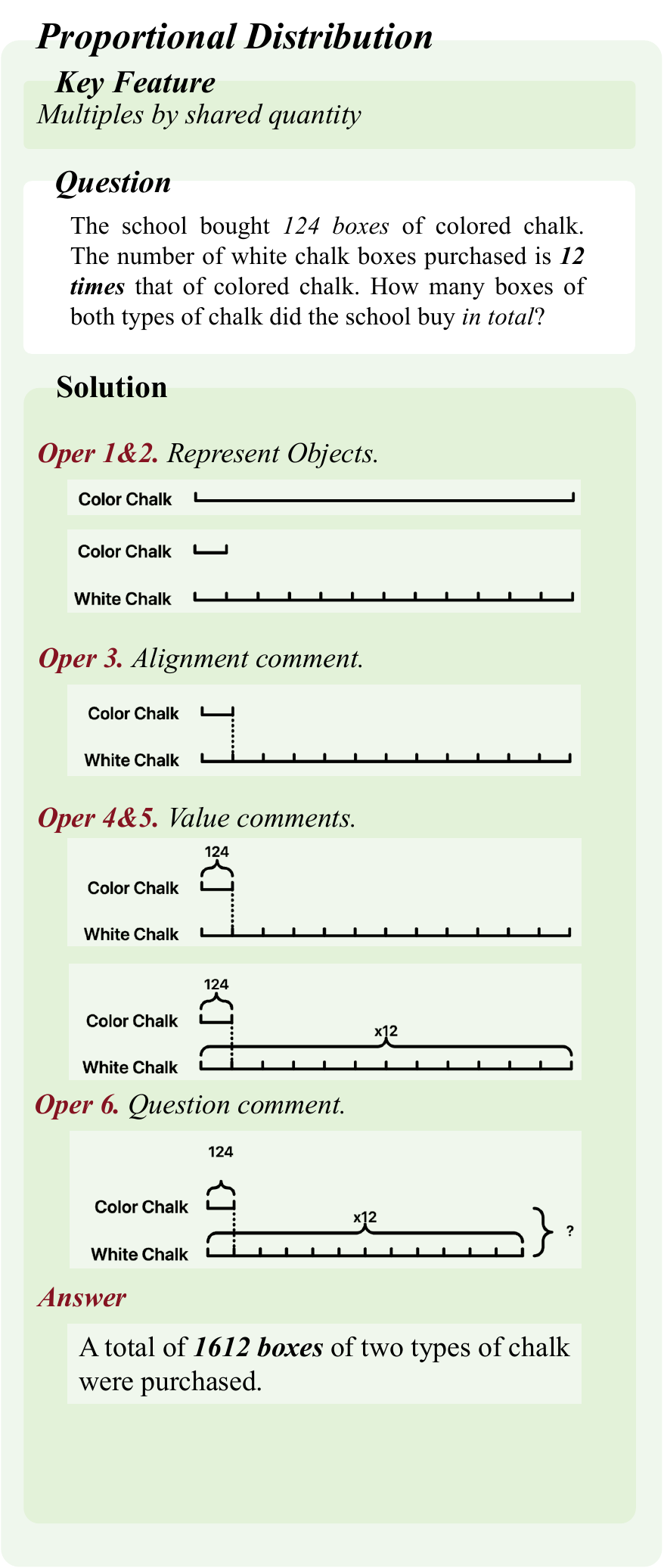}
  \caption{\textbf{Proportional Distribution.} The diagram enforces the multiplicative relation via repeated equal-length units and boundary alignment, and marks the queried total as an explicit unknown on the composed bar. TwD does not merely calculate $12\times 124$; instead, it enforces the multiplicative constraint via topological repetition. By rendering the `White Chalk' bar as a composite of 12 equal-length units aligned with the `Color Chalk' reference unit, the model transforms an abstract arithmetic operation into a concrete unit-repetition task, making the total sum visually deducible.}
  \label{fig:case_prop}
\end{figure}

\begin{figure*}[ht]
  \centering
  \includegraphics[width=0.95\textwidth,height=0.78\textheight,keepaspectratio]{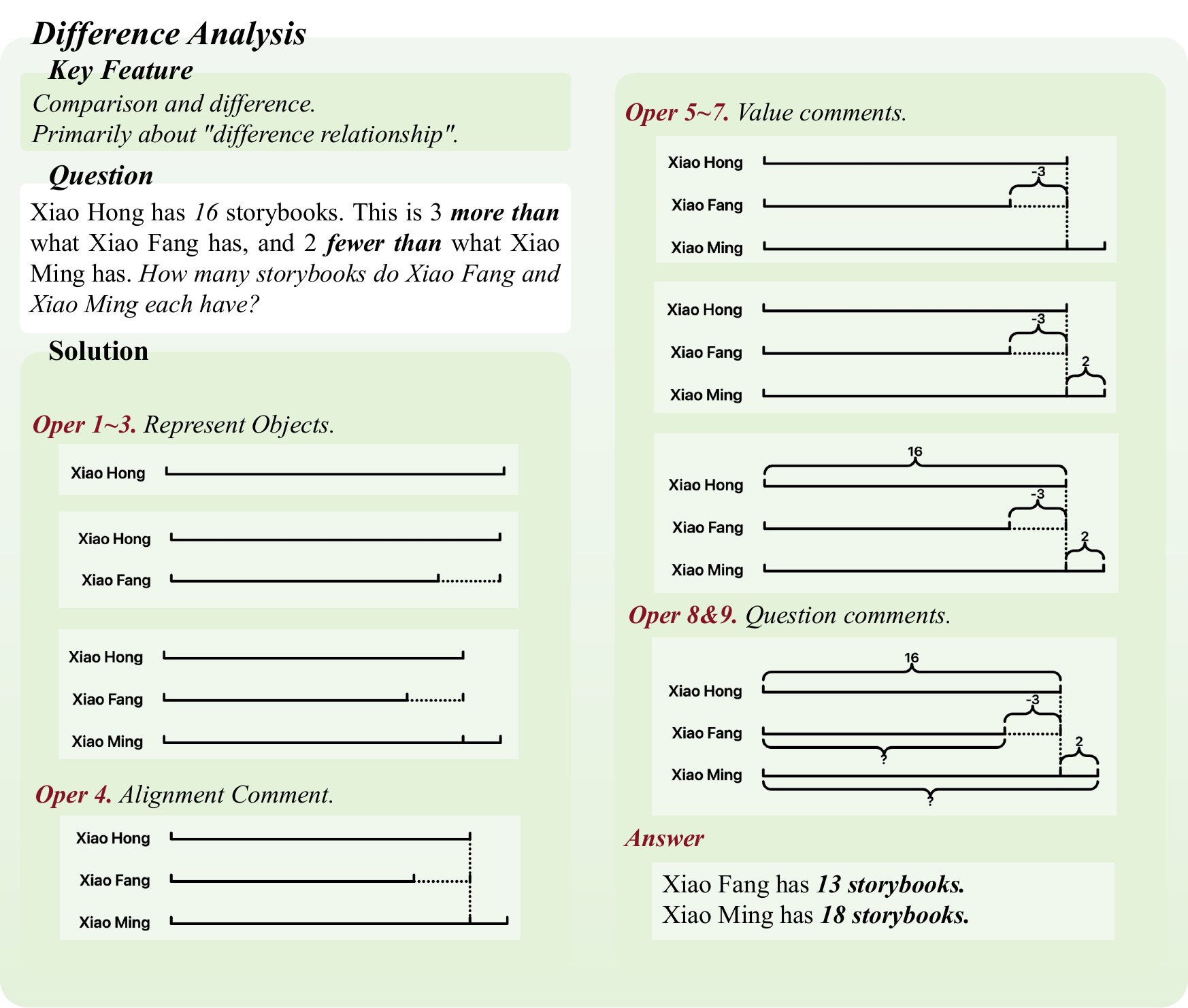}
  \caption{\textbf{Difference Analysis.} This example illustrates the Thinking with Drafting process on a multi-entity comparison problem. The model does not hallucinate the answer directly; instead, it performs logical reconstruction in steps: (1) instantiating objects (Oper 1-3), (2) enforcing topological alignment via vertical anchors (Oper 4), and (3) encoding ``more than/fewer than'' relations as explicit offset segments (Oper 5-7). This step-by-step grounding ensures that the final arithmetic inference is derived from a verified geometric structure.}
  \label{fig:case_diff}
\end{figure*}

\paragraph{Alignment-centric schemas.}
\textbf{Difference Analysis} (Figure~\ref{fig:case_diff}) and \textbf{Proportional Distribution} (Figure~\ref{fig:case_prop}) both require cross-object alignment to make relational constraints executable.
In Figure~\ref{fig:case_diff}, the DSL anchors one entity as a reference and encodes ``more than'' / ``fewer than'' relations as explicit offset segments, turning comparative language into a geometry-consistent subtraction layout.
In Figure~\ref{fig:case_prop}, the DSL realizes the multiplicative constraint by repeating equal-length unit segments and aligning boundaries across rows, so the ``$\times 12$'' relation is enforced by topology rather than inferred implicitly; the final query is then represented as a single unknown bracket on the composed total.

\begin{figure}[ht]
  \centering
  \includegraphics[height=0.78\textheight,keepaspectratio]{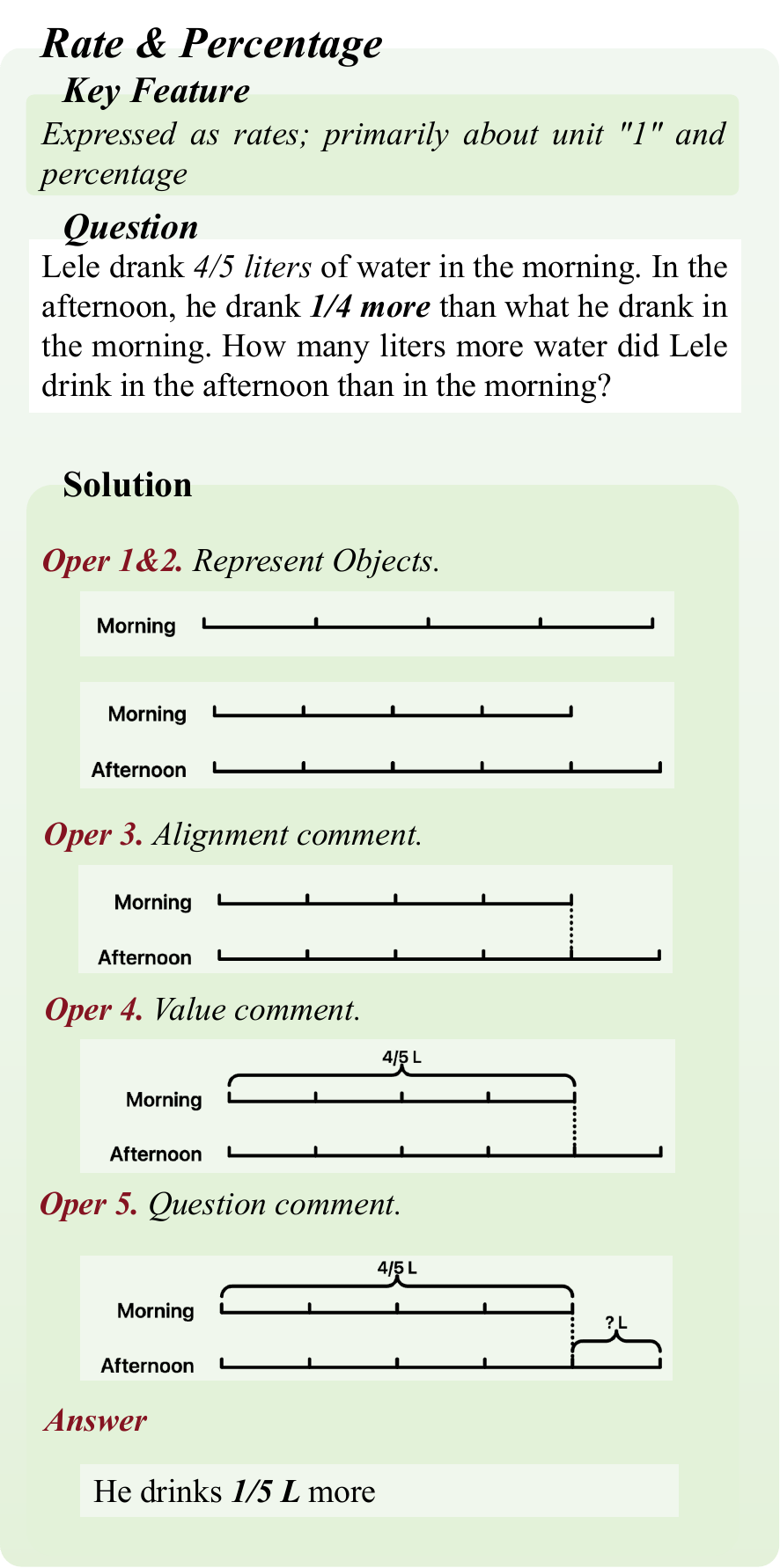}
  \caption{\textbf{Rate \& Percentage.} Fractional change is grounded by fixing the base as a unit reference and attaching the fractional increment as a dedicated subsegment aligned to the shared boundary. The model fixes the morning consumption as the holistic unit ``1'', and attaches the fractional increment ($1/4$) as a dedicated sub-segment aligned to the unit boundary. This explicit segmentation allows the model to visually isolate the $\Delta$ from the whole, preventing unit confusion.}
  \label{fig:case_rate}
\end{figure}

\begin{figure}[ht]
  \centering
  \includegraphics[height=0.78\textheight,keepaspectratio]{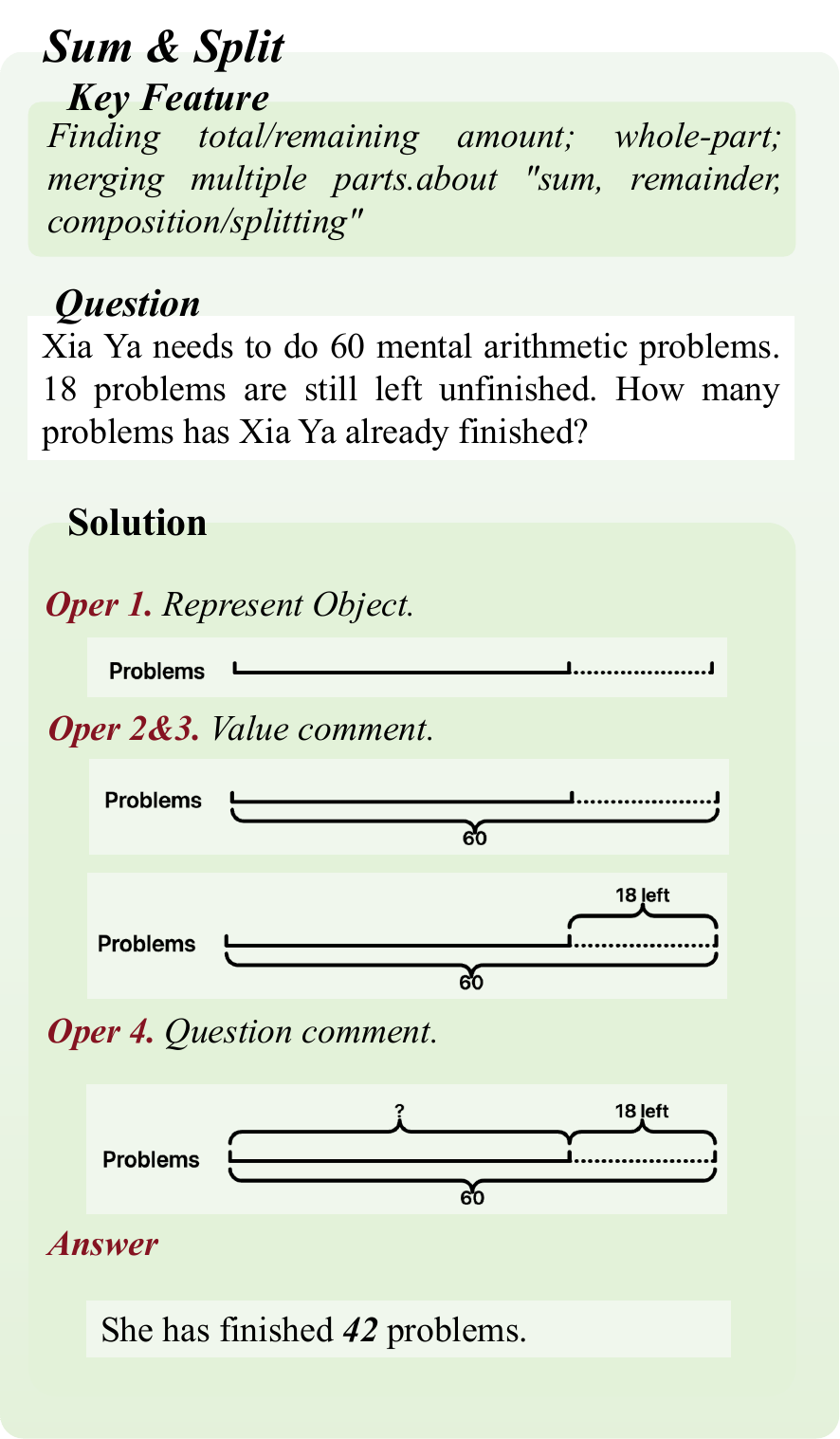}
  \caption{\textbf{Sum \& Split.} The whole--part structure is made explicit by isolating the known remainder segment and marking the target as the complementary unknown, directly supporting completion-by-subtraction. The unknown remainder is highlighted as the target, supporting solution by subtraction.}
  \label{fig:case_sum}
\end{figure}

\paragraph{Decomposition-centric schemas.}
\textbf{Sum \& Split} (Figure~\ref{fig:case_sum}) and \textbf{Rate \& Percentage} (Figure~\ref{fig:case_rate}) emphasize part--whole partition and unit grounding.
Figure~\ref{fig:case_sum} isolates the known remainder as a dedicated segment and marks the target as the complementary part, making the computation a direct completion on the bar.
Figure~\ref{fig:case_rate} grounds fractional change by first fixing the base quantity as the unit reference and then attaching the fractional increment as an explicit subsegment, reducing ambiguity about the comparison base.

\begin{figure}[ht]
  \centering
  \includegraphics[height=0.78\textheight,keepaspectratio]{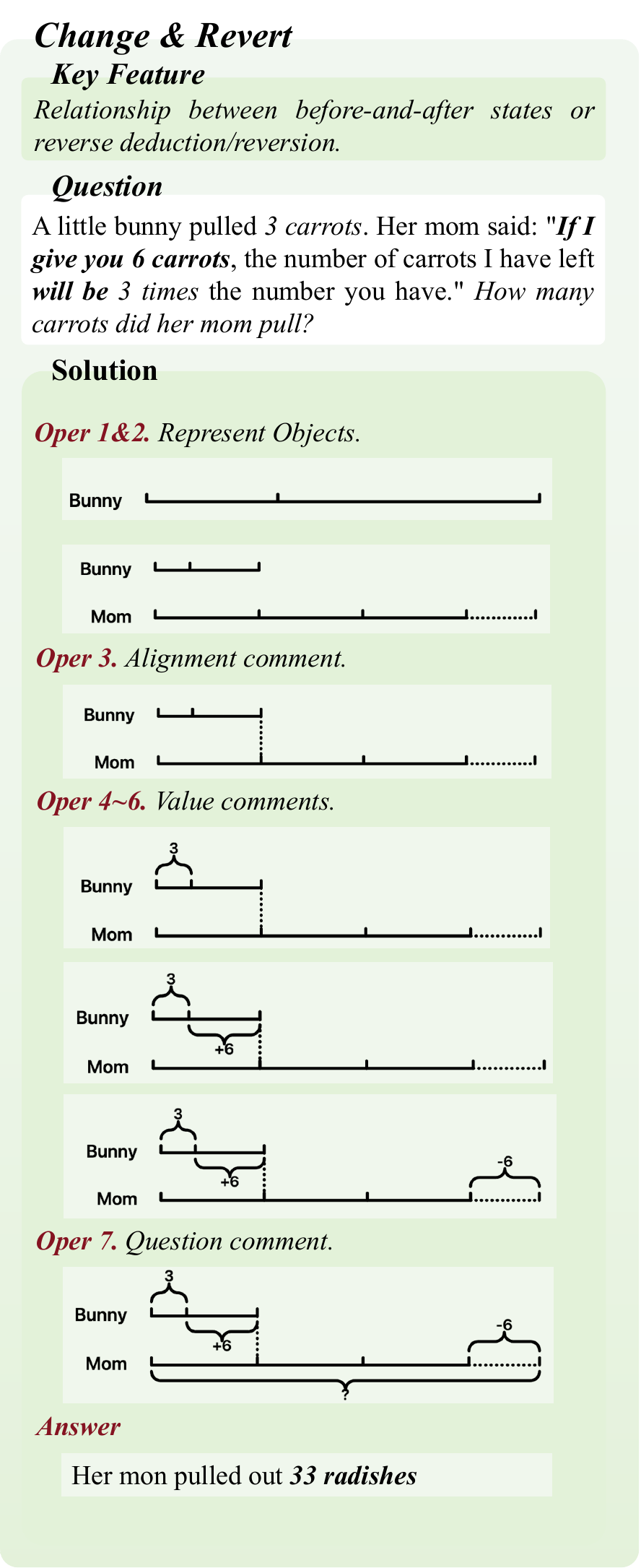}
  \caption{\textbf{Change \& Revert.} A counterfactual transfer is rendered as paired decrease/increase segments, after which the post-transfer constraint is imposed on the aligned after-state topology. This example illustrates how TwD handles hypothetical state transitions. The model employs a dual-segment representation where transfers are rendered as paired decrease/increase segments. Crucially, the ``post-transfer'' multiplicative constraint $\times 3$ is imposed not on the initial state, but on the aligned after-state topology. This proves the model's ability to reason about dynamic temporal states within a static spatial diagram.}
  \label{fig:case_change}
\end{figure}

\paragraph{State-transition schema.}
\textbf{Change \& Revert} (Figure~\ref{fig:case_change}) involves a counterfactual transfer and a post-transfer relation.
The DSL externalizes the hypothetical ``give'' operation with paired decrease/increase segments and then imposes the after-state constraint on the aligned configuration, enabling reverse deduction while keeping all visible labels faithful to the original statement (i.e., without leaking computed answers).

\begin{figure*}[ht]
  \centering
  \includegraphics[width=\linewidth]{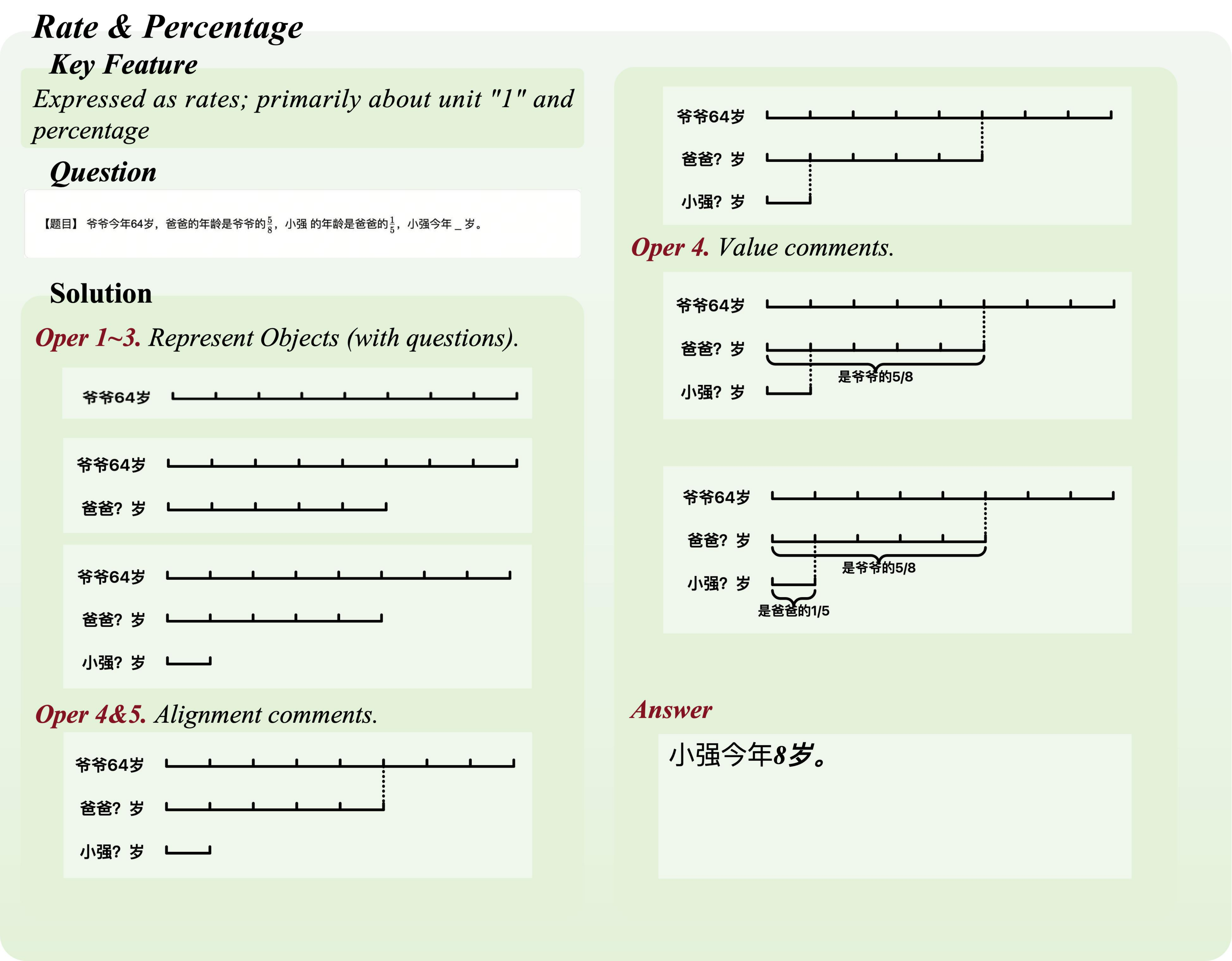}
    \caption{\textbf{Rate \& Percentage.} A complex multi-step ratio problem involving chain dependencies. TwD manages this hierarchy through cascading alignment: each subsequent row's length is topologically anchored to the specific fraction of the preceding row. The vertical dashed lines serve as transitive logic gates, ensuring that the final quantity is derived from a rigorously valid chain of geometric proportions, minimizing error propagation.}
  \label{fig:Rate-zh-1}
\end{figure*}

\begin{figure}[ht]
  \centering
  \includegraphics[height=0.78\textheight,keepaspectratio]{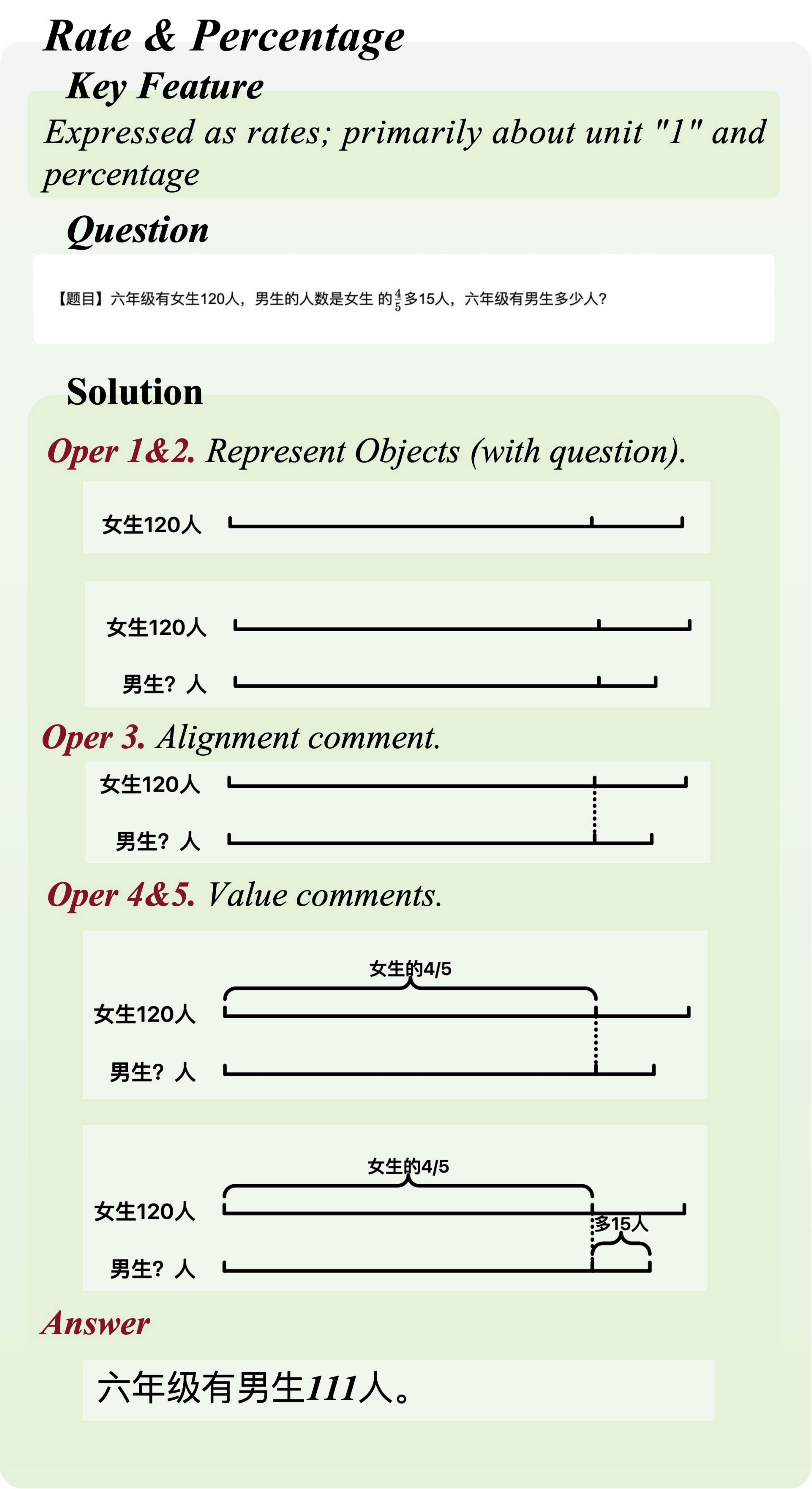}
    \caption{\textbf{Rate \& Percentage.} Fractional relationships are represented by fixing the base quantity as a unit reference and aligning proportional segments to this shared unit. The fractional change is visualized as a dedicated subsegment, supporting reasoning based on rates and percentages.}
  \label{fig:Rate-zh-2}
\end{figure}

\begin{figure}[ht]
  \centering
  \includegraphics[height=0.78\textheight,keepaspectratio]{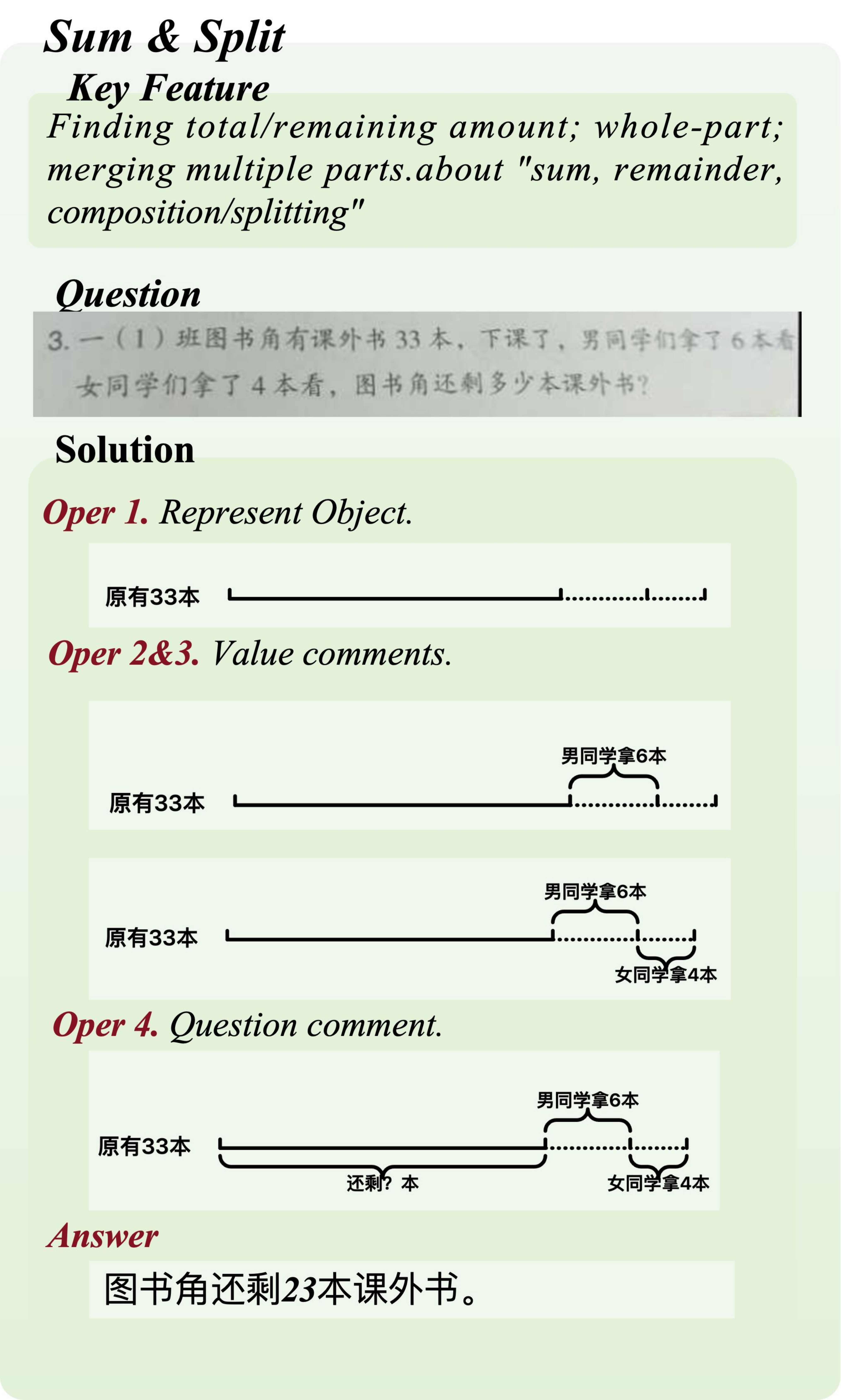}
    \caption{\textbf{Sum \& Split.} The whole--part structure is made explicit by isolating the known remainder segment and marking the target as the complementary unknown, directly supporting completion-by-subtraction. The unknown remainder is highlighted as the target, supporting solution by subtraction.}
  \label{fig:Sum-zh-2}
\end{figure}

\begin{figure}[ht]
  \centering
  \includegraphics[height=0.78\textheight,keepaspectratio]{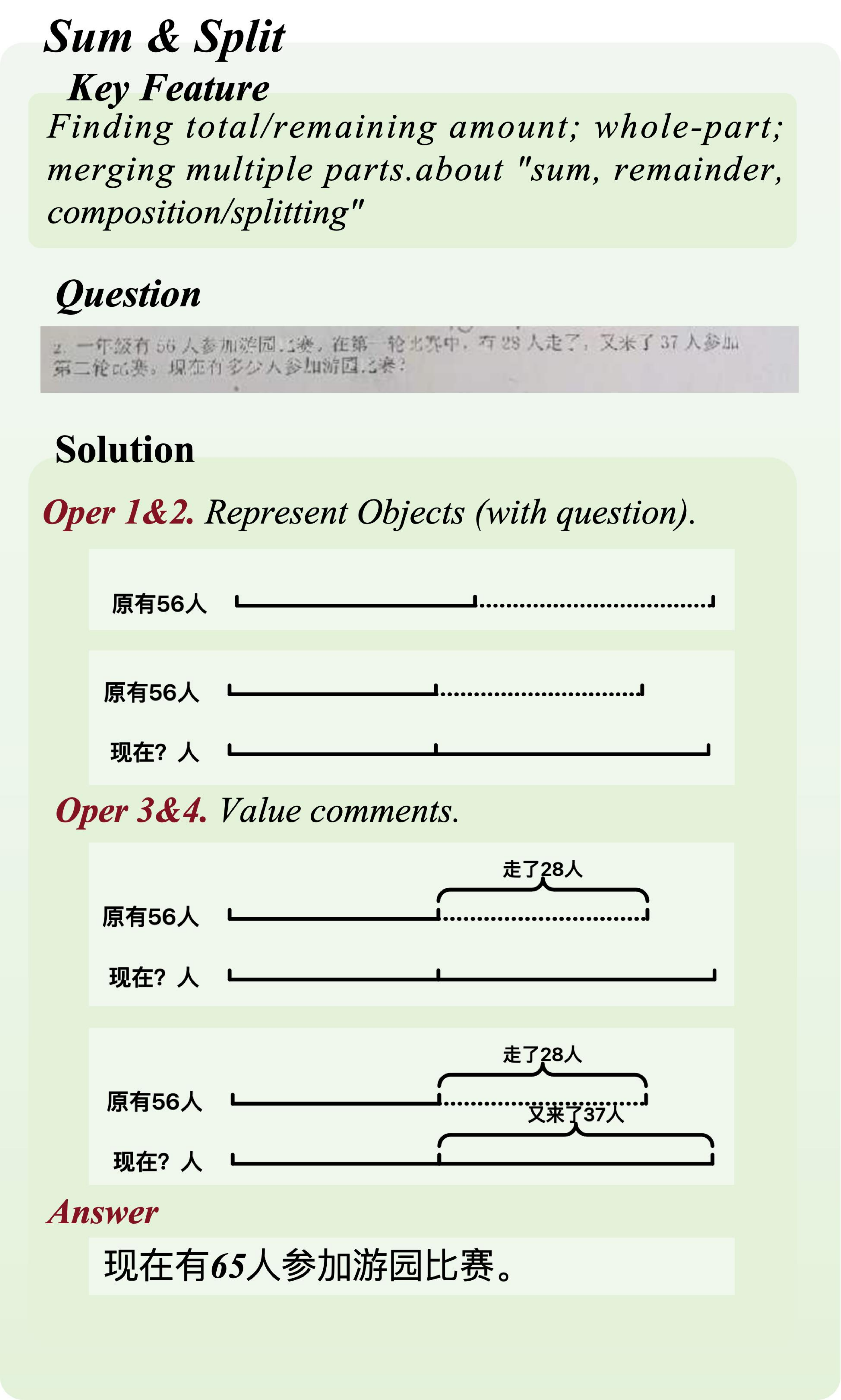}
    \caption{\textbf{Sum \& Split.} The whole--part structure is made explicit by isolating the known remainder segment and marking the target as the complementary unknown, directly supporting completion-by-subtraction. The final total is identified as the complementary unknown, supporting solution by subtraction and addition.}
  \label{fig:Sum-zh-3}
\end{figure}

\begin{figure*}[ht]
  \centering
  \includegraphics[width=\linewidth]{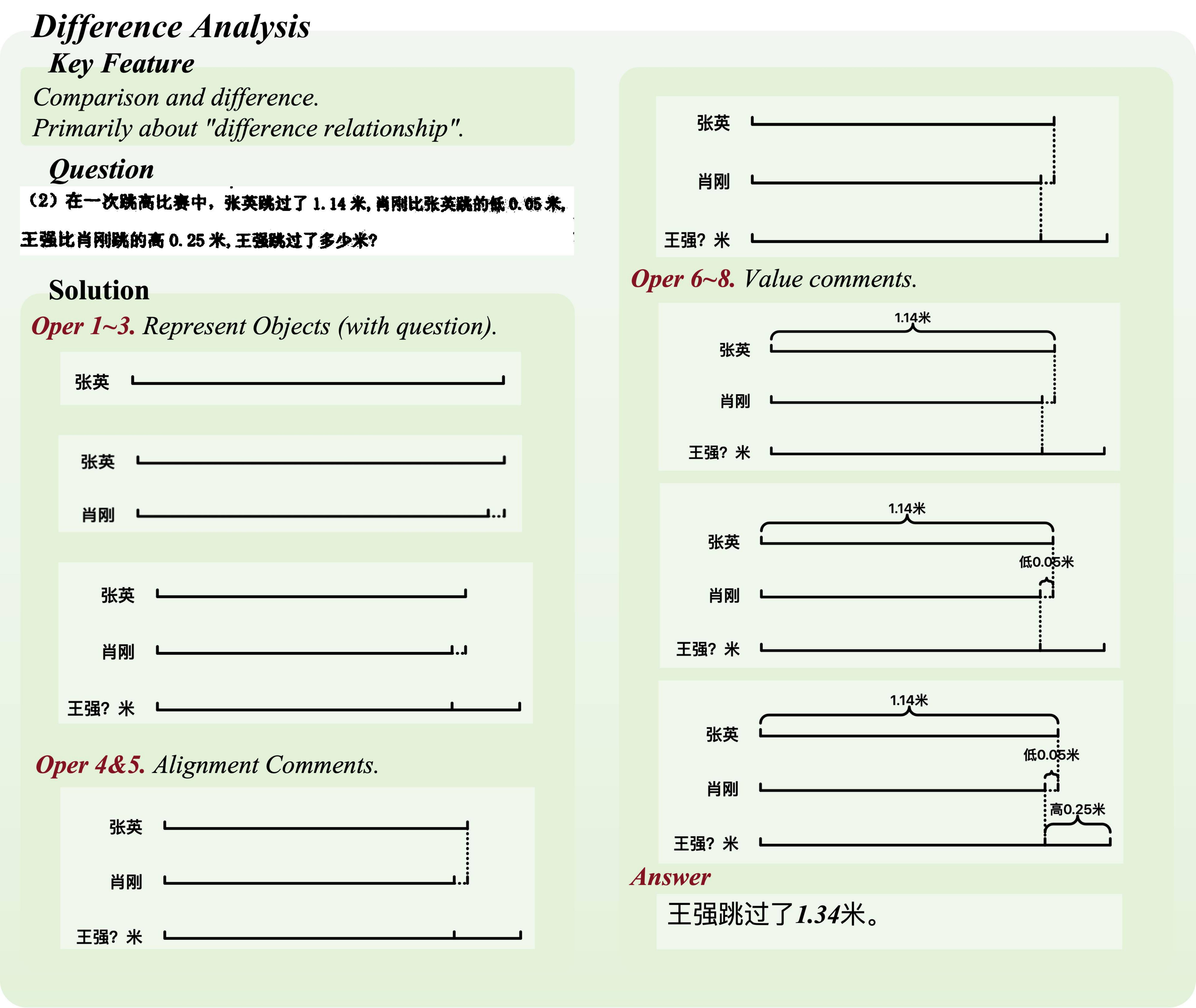}
    \caption{\textbf{Difference Analysis.} Application of TwD to a continuous-value scenario involving bidirectional differences (``lower than'' vs. ``higher than''). The system successfully decodes the textual constraints into precise spatial alignments. Note how the vertical dashed lines act as logical anchors, physically locking the relative positions of the reference entity and derived entities. This transforms an abstract arithmetic word problem into a concrete visual subtraction and addition task, mitigating logical errors in multi-step calculation.}
  \label{fig:Diff-zh-1}
\end{figure*}

\section{Additional Error Analysis}
\label{app:more_error}
We summarize a \textbf{Taxonomy of Structural Degeneration} observed in baseline diagrams, where the output may remain arithmetically compatible yet loses the structural invariants required for verification.

\subsection{Semantic Erasure: Multiplicative Topology Collapsed}
\begin{figure*}[t]
  \centering
  \includegraphics[width=\linewidth]{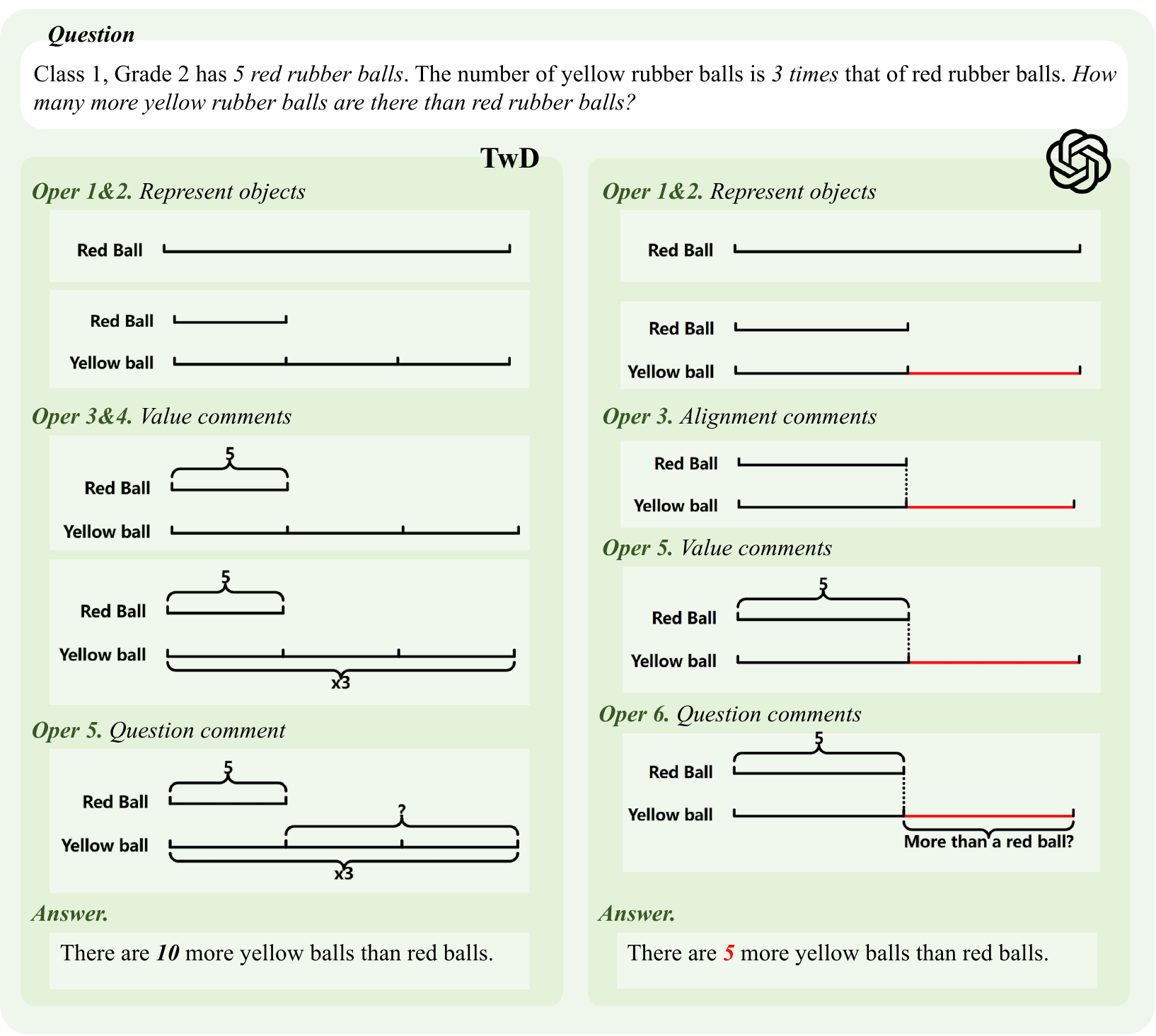}
    \caption{\textbf{Semantic Erasure.} The $\times 3$ constraint is collapsed into an additive layout, removing repeated-unit evidence. The baseline model suffers from Semantic Erasure: it collapses the multiplicative constraint into a generic additive layout, failing to render the repeated unit segments. TwD explicitly preserves the unit topology, rendering three distinct segments for the yellow ball row. This structural fidelity enforces the correct arithmetic operation.}
  \label{fig:error_balls}
\end{figure*}

In Figure~\ref{fig:error_balls}, the baseline collapses the given $\times 3$ constraint into an additive ``difference'' layout.
This erases the repeated-unit structure, so the multiplier is no longer visually provable even if the final arithmetic is correct.

\subsection{Label Injection: Numbers without Geometric Support}
\begin{figure*}[t]
  \centering
  \includegraphics[width=\linewidth]{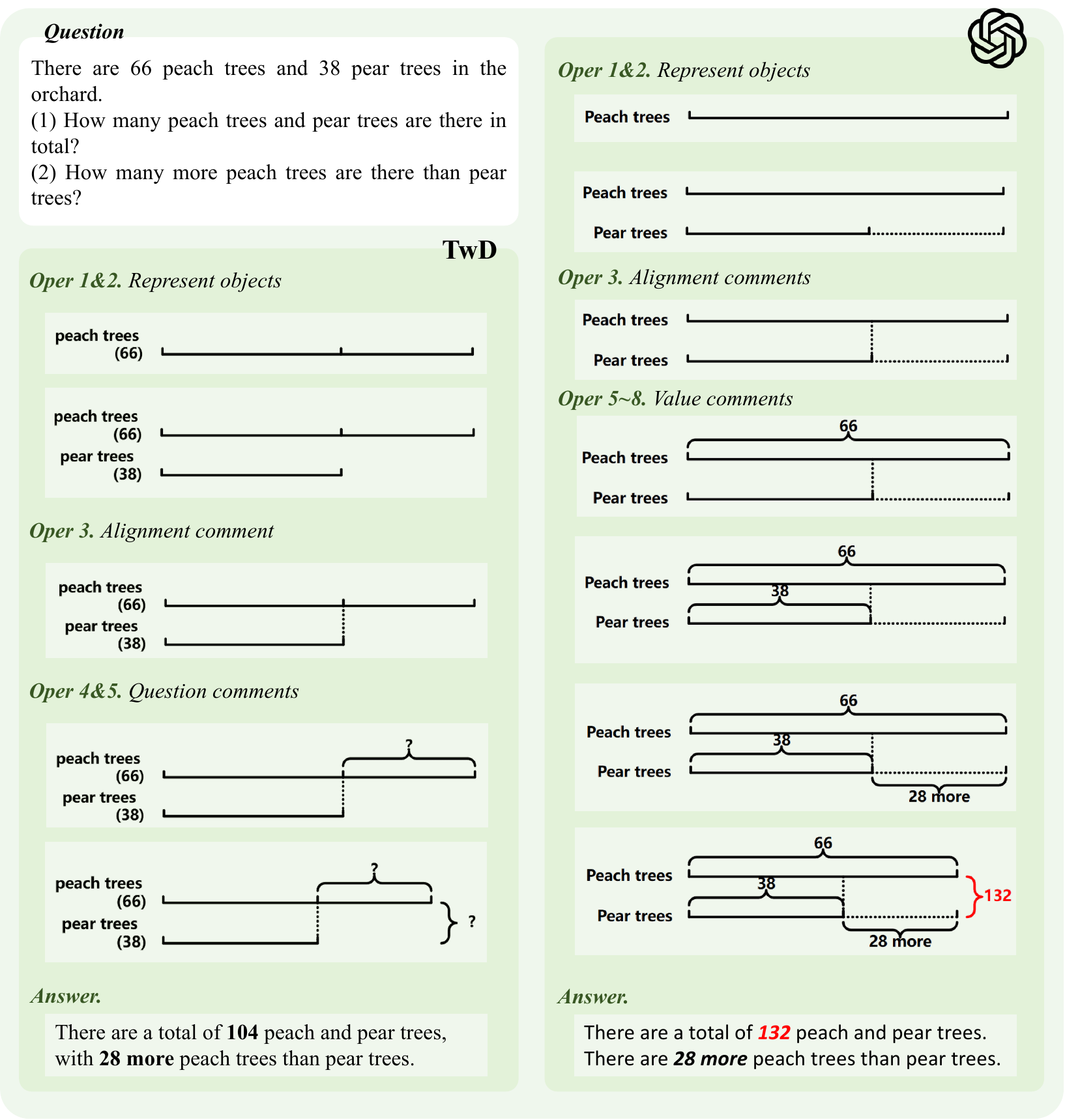}
    \caption{\textbf{Label Injection.} A computed value is written as text without a supporting sub-segment, yielding an ungrounded claim. The baseline model exhibits Label Injection: it hallucinates a computed value (``132'') and injects it as a text label without generating the supporting geometric sub-segments. The visual diagram thus becomes a deceptive artifact that does not physically represent the sum. TwD constructs the result bottom-up. By strictly aligning the start and end points of the `Peach' and `Pear' segments, it creates a valid geometric aggregation, ensuring the final answer is visually deducible.}
  \label{fig:error_trees}
\end{figure*}

Figure~\ref{fig:error_trees} shows a label--structure mismatch: the model writes the computed difference as text, but does not allocate a corresponding sub-segment.
The diagram therefore contains \emph{claims} without geometric evidence, and downstream reasoning can mistakenly treat labels as quantities.

\subsection{Alignment Conflict: Incompatible Global Boundaries}
\begin{figure*}[t]
  \centering
  \includegraphics[width=\linewidth]{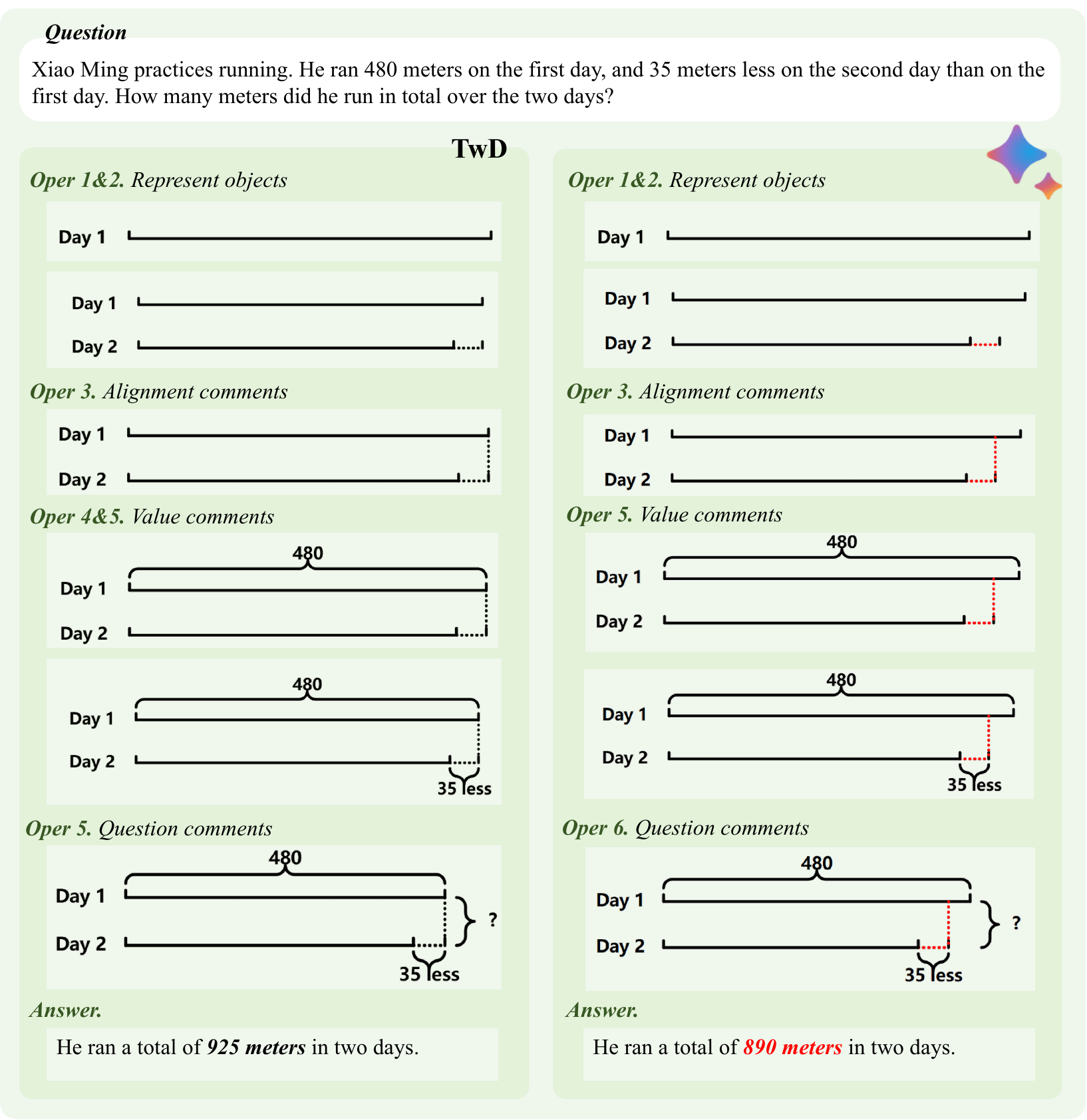}
    \caption{\textbf{Alignment Conflict.} Conflicting global boundaries break the stability of cross-row relations. The baseline model generates an Alignment Conflict: the vertical dashed line (alignment anchor) is misplaced, visually suggesting that Day 2 is longer than Day 1 despite the label ``35 less''. This topological contradiction breaks the logical chain, leading to an erroneous calculation. TwD correctly places the subtractive anchor. The dashed line precisely demarcates the difference segment, enforcing a consistent spatial logic where the length of Day 2 is physically constrained to be shorter, guiding the correct subtraction.}
  \label{fig:error_run}
\end{figure*}

In Figure~\ref{fig:error_run}, the baseline mixes incompatible alignment cues: dashed completion implies one shared endpoint, while vertical guides declare another boundary.
This breaks global boundary consistency, so the ``less by 35'' relation is not stably encoded in the diagram.

\section{Potential Risks}

We identify two primary risks that stem from the formalization of reasoning introduced by the Thinking with Drafting (TwD) paradigm, particularly in educational contexts.

First, the use of a structured DSL may amplify \textit{automation bias}.
Because the generated diagrams resemble formal proofs, users may conflate structural validity with semantic correctness, implicitly assuming that a well-formed intermediate representation guarantees a correct solution.

Second, TwD introduces a risk of \textit{cognitive offloading} that may lead to \textit{skill atrophy} in diagrammatic reasoning.
By externalizing key steps of problem decomposition and visualization, the system may reduce the learner's engagement in constructing and maintaining structural invariants.
Over time, excessive reliance on automated drafting can weaken the user's ability to independently translate textual constraints into spatial representations, undermining the development of foundational visual reasoning skills.

\end{document}